\documentclass[journal,twoside,web]{ieeecolor}
\usepackage{generic}
\usepackage{cite}
\usepackage{amsmath,amssymb,amsfonts}
\usepackage{algorithmic}
\usepackage{graphicx}
\usepackage{algorithm,algorithmic}
\usepackage{hyperref}
\hypersetup{hidelinks}
\usepackage{textcomp}
\usepackage{epstopdf}
\usepackage{graphicx}
\DeclareGraphicsExtensions{.pdf,.png,.jpg}

\usepackage{multirow}
\usepackage[margin=1in]{geometry}

\usepackage{subcaption}

\usepackage{titlesec}
\titlespacing*{\section}{0pt}{6pt}{3pt}
\titlespacing*{\subsection}{0pt}{4pt}{0pt}

\providecommand{\refname}{References}

\def\BibTeX{{\rm B\kern-.05em{\sc i\kern-.025em b}\kern-.08em
    T\kern-.1667em\lower.7ex\hbox{E}\kern-.125emX}}
\usepackage[font=small]{caption}

\begin{document}
\title{fMRI-TAMCL: Text-Anchored Supervised Multimodal Contrastive Learning for fMRI-Based Brain Disorder Classification}
\author{Juliana Mantebea Danso, \IEEEmembership{Student Member, IEEE}, Enoch Opanin Gyamfi, and Mylene C.Q.  Farias, \IEEEmembership{Senior Member, IEEE}
% \thanks{This paragraph of the first footnote will contain the date on which you submitted your paper for review. It will also contain support information, including sponsor and financial support acknowledgment. For example, ``This work was supported in part by the U.S. Department of Commerce under Grant 123456.'' }
% \thanks{The next few paragraphs should contain the authors' current affiliations, including current address and e-mail. For example, First A. Author is with the National Institute of Standards and Technology, Boulder, CO 80305 USA (e-mail: author@boulder.nist.gov). }
\thanks{J.M. Danso and M.Q. Farias are  with the Department of Computer Science, Texas State University, San Marcos, TX 78666 USA (e-mail:  juliana@txstate.edu and mylene@ieee.org).}
\thanks{E. O. Gyamfi is with the University of Technology and Applied Sciences, Navrongo, Ghana (email:eogyamfi@utas.edu.gh).}
% }
}
\maketitle

\begin{abstract}
Resting-state fMRI is important in the classification of brain disorders, but highly multimodal and exhibits strong multisite heterogeneity. Existing methods fuse images, BOLD-based functional connectivity, and phenotypic data modalities. Unlike other medical imaging datasets, rs-fMRI datasets rarely include a text modality, so they are generated from phenotypic data or BOLD activations. These text generation methods rely on fixed assumptions for subjects, sites, devices, and protocols, leading to poor generalization across datasets. We propose fMRI-TAMCL, a text-anchored multimodal contrastive learning framework that integrates fMRI images, sparse FC, and generated subject-specific text. Its Subject-Adaptive Threshold Derivation module generates BOLD activation text, while Feature–Value Serialization module generates phenotypic text. All three modalities are encoded as clustered graphs, projected onto a shared unit hypersphere  space, aligned using pairwise, text-anchored supervised contrastive learning, and fused with attention. fMRI-TAMCL proves its generalization capability across five datasets outperforming 29 baselines with 78.6\%–86.4\% accuracy in downstream classification.
\end{abstract}


\section{Introduction}
\label{sec:introduction}
\IEEEPARstart{R}{esting}-state functional magnetic resonance imaging (rs-fMRI) enables automated large-scale classification of brain disorders, but its use is limited by multi-site heterogeneity and multimodality. Public rs-fMRI datasets are collected across sites, scanner manufacturers, imaging protocols, and cohorts, that complicate generalizing model assumptions. Typical rs-fMRI studies include multiple heterogeneous modalities: images, functional connectivity matrices (FC), and phenotypic records tabulating demographic, cognitive, behavioral, and other clinical data. Consequently, many rs-fMRI diagnostic models are modality-specific: CNNs for images, GNNs for FCs, and conventional ML for tabular phenotypic data. This unimodal approach limits the use of complementary cross-modal information. Among them, FC matrices are especially common. For instance, BrainGNN \cite{li2021braingnn} learns interpretable region-of-interest (ROI) and community-level biomarkers from FC-derived graphs. Recent multimodal strategies tend to fuse FCs with other quantitative neuroimaging modalities (MRI, PET, DTI, EEG, and image) \cite{yen2023exploring} and largely ignore text data.

A critical gap in therefore created in integrating clinical text and neuroimaging data, reducing the field’s opportunity to leverage recent advances in large language models (LLMs). Contrastive learning, a common multimodal learning strategy, also offers an opportunity to fuse neuroimaging and text by aligning them in a shared embedding space, while supervised contrastive learning further improves diagnostic class separation through label-defined positive pairs \cite{yuan2025survey}. These opportunities have recently motivated text-enhanced rs-fMRI diagnosis, but because such datasets rarely contain subject-level text data, existing methods must derive text representations from brain activations. These existing methods typically use fixed activation thresholds, mapping ROI-level BOLD activations to labels (e.g., weak, moderate, strong). They assume similar activation distributions across subjects and sites. However, in multisite datasets, a fixed threshold cannot label different neural patterns due to inter-subject and inter-site variability in scanner and acquisition protocols. Fixed thresholds therefore introduce systematic label noise into the text modality, degrading downstream multimodal fusion. Fixed thresholds are also difficult to calibrate across datasets, resulting in poor generalizability across different datasets and brain disorders. \cite{dai2022multimodal}.

We present fMRI-TAMCL, a text-anchored multimodal contrastive learning framework for rs-fMRI–based brain disorder diagnosis that integrates three subject-level modalities: fMRI images, sparse FC, and generated text data. The Subject-Adaptive Threshold Derivation (SATD) module models each subject’s BOLD activation distribution with a Gamma–Gaussian mixture model (GGMM) and derives subject-level activation thresholds from the weighted
component intersections, replacing fixed thresholds. The Feature-Value Serialization (FVSerial) module converts tabular phenotypic data also into text data. SATD+FVSerial text data are merged into structured neurobiological text data that act as semantic anchors for contrastive learning. All modalities are represented as graphs via a unified construction pipeline, modularity-based clustering, and a GNN for feature embeddings. The image, FC and text embeddings are projected onto a shared unit hypersphere  latent space and optimized with three contrastive objectives: pairwise cross-modal alignment, text-anchored alignment, and supervised diagnostic class separation. An attention-based fusion module assigns subject-specific modality weights prior to classification.

The main contributions of this study are: (1) SATD, a subject-adaptive GGMM-based thresholding framework that generates text data from BOLD activation, and FVSerial, which also generates text data from tabular phenotypic data. These generated text data are unified to a single semantic textual modality; (2) a graph-based multimodal feature extraction pipeline for images, sparse FCs, and the generated texts; (3) a text-anchored, supervised multimodal contrastive learning with multi-head self-attention fusion for diagnostic classification; and (4) extensive experiments on five multisite datasets evaluating the generalization capability of the proposed model on diagnostic performance, robustness (with ablations), sensitivity, computational efficiency, and interpretability.

\section{Proposed Methodology}
\label{sec:Methodology}
\begin{figure*}[htbp!]
    \centering
    \includegraphics[width=0.90\textwidth]{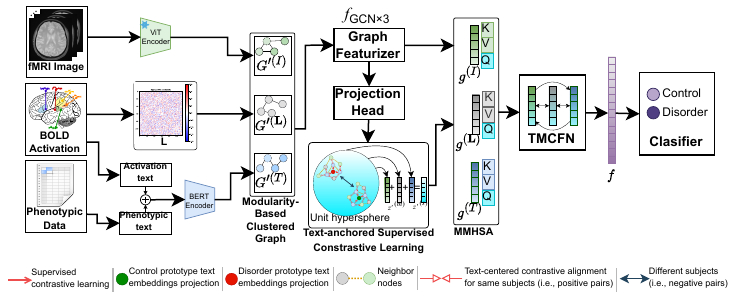}
    \caption{Overview of the proposed fMRI-TAMCL framework. Multimodal fMRI image, FC, and text features are encoded, projected into a shared contrastive latent space, fused through attention, and classified for diagnosis}
    \label{Figure1_Overview_fMRI-TAMCL}
\end{figure*}

% \subsection{Overview}
The fMRI-TAMCL architecture in Figure \ref{Figure1_Overview_fMRI-TAMCL} has two stages: (1) graph-based feature extraction and (2) contrastive learning-Based diagnosis. Next, we describe these stages.
% Image, sparse FC, and text modalities are encoded as graph-structured embeddings, projected into a shared unit-hyperspherical latent space, aligned by text-anchored supervised contrastive learning, fused via attention, and passed to a classification module for diagnosis.

%[width=.9\textwidth]

\subsection{Stage 1: Modality-Specific Feature Extraction} 
We build separate graph-based feature extractors for the image, sparse FC, and text modalities, \(m \in \{I, \mathbf{L}, T\}\). For the image branch, the fMRI volume \(I\) is linearly embedded to \(I_e\), then encoded by a ViT into patch features \(F^{(I)} = \{f_1^{(I)}, \dots, f_{d_I}^{(I)}\}\). Each patch is a graph node, with edges given by a \(K\)-nearest-neighbor graph under cosine similarity:
\[
S_{ij}^{(I)}=\frac{f_i^{(I)\top} f_j^{(I)}}{\lVert f_i^{(I)}\rVert \,\lVert f_j^{(I)}\rVert}.
\]
This defines the image graph \(G^{(I)}\), which is coarsened to \(G'^{(I)}\) via modularity-based clustering. A graph featurization module then encodes \(G'^{(I)}\) into the image embedding \(g^{(I)}\).

In the FC feature extraction branch, raw BOLD time series are converted into a sparse FC matrix $\mathbf{L}$ using the Log-regularized Laplacian Sparse Functional Connectivity (LLSFC) method \cite{danso2026ensemble}. LLSFC estimates the weight matrix $\mathbf{W}$ by minimizing
\begin{equation}
    R(\mathbf{W}) = -\alpha [\cdot\ \mathbf{1}]^{T} \log(\mathbf{W} \cdot \mathbf{1}) + \beta \| \mathbf{W} \|_{F}^{2},
\end{equation}
where $\alpha$ and $\beta$ are regularization hyperparameters, $\mathbf{1}$ is the all-ones vector, and $\|\cdot\|_{F}$ is the Frobenius norm. The log-barrier term promotes non-degenerate connectivity, while the Frobenius term prevents the FC matrix from becoming overly dense and dominating the model. As in the image branch, $\mathbf{L}$ defines a graph $G^{(L)}$ whose edges encode semantic similarity. A modularity-based clustering coarsen this graph into $G'^{(L)}$ and, in parallel, a graph featurizer produces the sparse FC graph embedding $g^{(L)}$.

In the text feature extraction branch, raw tabular phenotypic and brain activation data are converted into textual descriptions $T_{pheno}$ and $T_{brain}$, whose concatenation $T_{pheno} \oplus T_{brain}$ forms $T_{fMRI}$. An LLM refines $T_{fMRI}$ into $T_{refined}$, which a generic BERT model encodes into contextual token embeddings $F^{(T)}$. Then again, each token embedding $f^{(T)}_{i}$ becomes a node in graph $G^{(T)}$, with edges defined by semantic similarity scores $S_{ij}^{(T)}$. A modularity-based clustering algorithm merges redundant semantic nodes, yielding a coarsened graph $G'^{(T)}$, which a three-layer GCN featurizer processes to produce the text graph embedding $g^{(T)}$.

\begin{figure}[!tb]
    \centering
    \includegraphics[width=0.9\columnwidth]{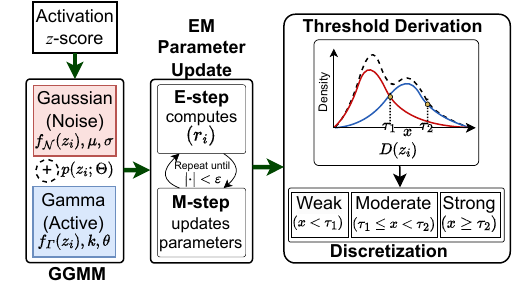}
    \caption{Subject-Adaptive Ternary Discretization (SATD) method for thresholding ROI-level fMRI timeseries data}
    \label{Figure2_SATD}
\end{figure}

% \subsection{Generating Brain Text Descriptions}
After outlining the graph-processing pipeline, we focus on the text-generation module that produces subject-level textual representations. We first describe the generation of the brain-activation description \(T_{brain}\), followed by the phenotypic text description \(T_{pheno}\). The SATD module (Figure \ref{Figure2_SATD}) converts raw BOLD timeseries into the brain activation text representation \(T_{brain}\). It uses GGMM to adaptively discretize ROI-level activations into natural-language labels \textit{“weak”}, \textit{“moderate”}, and \textit{“strong”}. For each subject \(i\), \(z\)-score normalization is applied to the atlas-defined \(N\) ROI BOLD timeseries, producing an activation vector \(z_i \in \mathbb{R}^N\). ROI activations follow a mixture of Gaussian noise \(f_G(z_i)\) for resting or suppressed regions and a Gamma process \(f_\Gamma(z_i)\) for elevated responses:
\[
    p(z_i; \Theta) = (1 - \pi)\,\underbrace{\mathcal{N}(z_i; \mu, \sigma^2)}_{f_G(z_i)} 
    + \pi\,\underbrace{\Gamma(z_i; k, \theta)}_{f_\Gamma(z_i)},
\]
where \(\Theta = \{\pi, \mu, \sigma, k, \theta\}\), \(\pi \in (0,1)\) is the mixture weight of the active (Gamma) component, and \(\mu \approx 0\). The Gaussian term models near-zero noise, and the Gamma term captures right-skewed activation, with shape \(k\) setting skewness and scale \(\theta\) setting dispersion.
% \[
%     f_G(z_i) = (2\pi\sigma^2)^{-\frac{1}{2}} \exp\!\left(-\frac{(z_i-\mu)^2}{2\sigma^2}\right)
% \]

% \[
%     f_\Gamma(z_i) = \frac{z_i^{k-1} \exp(-z_i/\theta)}{\theta^k \Gamma(k)}
% \]

% \subsubsection{Parameter Estimation via Expectation-Maximization (EM)}
In the expectation step (E-step in Figure \ref{Figure2_SATD}), the posterior responsibility of ROI for \(i\) is the active mixture component at iteration \(t\) computed as
\[
r_i^{(t)} =
\frac{f_\Gamma^{(t)}(z_i)}
{p(z_i;\Theta^{(t)})}.
\]
In the maximization step (M-step in Figure \ref{Figure2_SATD}), the mixture weight and Gaussian parameters are updated as
\begin{equation}
\begin{aligned}
\pi^{(t+1)}
&=
\frac{1}{N}
\sum_{i=1}^{N} r_i^{(t)},
\\[3pt]
\mu^{(t+1)}
&=
\frac{
\sum_{i=1}^{N}
\bigl(1-r_i^{(t)}\bigr)x_i
}{
\sum_{i=1}^{N}
\bigl(1-r_i^{(t)}\bigr)
},
\\[3pt]
\sigma^{2(t+1)}
&=
\frac{
\sum_{i=1}^{N}
\bigl(1-r_i^{(t)}\bigr)
\bigl(x_i-\mu^{(t+1)}\bigr)^2
}{
\sum_{i=1}^{N}
\bigl(1-r_i^{(t)}\bigr)
}.
\end{aligned}
\label{eq:em_updates}
\end{equation}
These are the standard EM updates for the mixture proportion and the Gaussian mean and variance, using \(1 - r_i^{(t)}\) as the effective Gaussian weights. For the active component, we define the weighted statistics
\[
\hat{\mu}_a = \frac{\sum_{i=1}^N r_i z_i}{\sum_{i=1}^N r_i},
\quad
\hat{L}_a = \frac{\sum_{i=1}^N r_i \ln z_i}{\sum_{i=1}^N r_i},
\]
and the sufficient statistic
\[
s = \ln(\hat{\mu}_a) - \hat{L}_a.
\]
The Gamma parameters at iteration \((t+1)\) are then
\[
k^{(t+1)} =
\frac{3 - s + \sqrt{(s - 3)^2 + 24s}}{12s},
\quad
\theta^{(t+1)} = \frac{\hat{\mu}_a}{k^{(t+1)}}.
\]
We iterate until the log-likelihood satisfies the convergence check
\(\lvert \ell^{(t+1)} - \ell^{(t)} \rvert < \varepsilon\). To avoid local optima, we run $n_{init}=5$ independent random restarts, keeping the GGMM parameters with the highest final log-likelihood.

% \subsubsection{Adaptive Threshold Derivation and Discretization}

After estimating the optimal GGMM parameters $\widehat{\Theta} = \{\widehat{\pi}, \widehat{\mu}, \widehat{\sigma}, \widehat{k}, \widehat{\theta}\}$, subject-specific thresholds are given by the intersections between the fitted Gaussian and Gamma components. To find them, we define the inter-component difference function
\begin{equation}
    D({z_i}) = (1 - \widehat{\pi}) \cdot \mathcal{N}(z_i; \widehat{\mu}, \widehat{\sigma}^2) 
           - \widehat{\pi} \cdot \Gamma(z_i; \widehat{k}, \widehat{\theta}),
\end{equation}

Adaptive thresholds are the roots of $D({z_i})$, denoted as $x$. $D({z_i})=0$ when the posterior probabilities of noise and active components are equal and overlap. Representing $\tau_1 < \tau_2$ as these thresholds, we map ROI activation levels to
\begin{equation}
    f(x) = \begin{cases} 
      \text{weak}, & x < \tau_1, \\
      \text{moderate}, & \tau_1 \le x \le \tau_2, \\
      \text{strong}, & x \ge \tau_2.
   \end{cases}
\end{equation}
Weak intensities are dominated by

Gaussian (noise) intensities are labeled \textit{``weak''}. $D({z_i})=0$ cases are labeled \textit{``moderate''}. Gamma (active) intensities are labeled \textit{``strong''}. If the EM algorithm does not converge, thresholds at the 33rd and 67th percentiles are used. If only one intersection is found, it is used as the lower threshold $\tau_1$, and the upper threshold is set to the theoretical Gamma mean, $\tau_2 = 1.5\,\widehat{k}\,\widehat{\theta}$.

% \subsubsection{Converting Tabular Phenotypic Data to Phenotypic Text}

Next, we generate $T_{pheno}$ using serialization of the feature-value pair (FVSerial). This converts each subject's raw tabular phenotypic data into a representation more naturally interpretable by an LLM. LLMs generally handle natural-language text more effectively than raw tabular data. Phenotypic data has rows as subjects' record values and columns as their features. These columns often use abbreviated or \texttt{\small CAPITAL\_SNAKE\_CASE} names, making them lack clear semantics and suboptimal for LLM processing. Our FVSerial module follows the TabLM framework \cite{hegselmann2023tabllm} to convert raw feature-value (column-row) pairs such as \texttt{\small SUB\_ID = 50002, AGE\_AT\_SCAN = 16.77, SEX = Male, FIQ = 103} in ABIDE, to human-readable feature-value pairs in sentence templates such as ``\texttt{\small The Subject ID is 50002.}'', ``\texttt{\small The Age is 16.77.}'', ``\texttt{\small The Sex is Male.}'', and ``\texttt{\small The Full-Scale Intelligence Quotient is 103.}''. These sentences follow the table column order. We then use a T0 encoder–decoder LLM model, a T5-based architecture trained on diverse prompt-oriented task, to independently convert each template-instantiated sentence into a concise natural-language text with relevant clinical context. T0 is finetuned on a table-to-text generation task\footnote{\url{https://huggingface.co/Narrativaai/bloom-560m-finetuned-totto-table-to-text}}.

% In the following, we describe the clustering and graph-featurization modules that are shared across all three modality-specific, graph-based feature extractors.

% \begin{figure*}[!tb]
%     \centering
%     \includegraphics[width=2.0\columnwidth]{Figure3_Converting_Pheno_Text.pdf}
%     \caption{Converting Tabular Phenotypic Data to Phenotypic Text}
%     \label{figure3_Converting_Pheno_Text}
% \end{figure*} 

% \subsection{Modularity-Based Graph Clustering.}
After describing the text-generation module, we revisit the graph-processing pipeline and explain how the generated text, image, and sparse FC graph representations are used to complete the downstream task. Because image, FC, and text graphs, denoted $G^{(I)}, G^{(L)}$, and $G^{(T)}$, are built from pairwise node similarity, they contain many semantically coherent communities. This can make each graph contain redundant nodes with overlapping or similar information. Direct multimodal fusion at the node-level is therefore suboptimal. To address this, we apply modularity-based clustering to each graph, grouping nodes into higher-level semantically meaningful clusters. This produces a soft cluster assignment matrix $\mathbf{C} \in \mathbb{R}^{n \times c}$, where each row gives the probability that a node belongs to each cluster, thus reducing the semantic redundancy at the node-level.

Since graphs are built from node similarity rather than random generation, a clustering algorithm may confuse a meaningful semantic community with a region of high edge density that appears by random. The modularity matrix $\mathbf{C}$ addresses this by testing whether a connection between two nodes reflects substantive similarity or is likely random. Formally, it subtracts a random-graph baseline from the observed adjacency structure:
\begin{equation}
    \mathbf{B} = \mathbf{A} - \frac{d d^T}{2e},
\end{equation}
where $\mathbf{A}$ is the adjacency matrix, $d$ is the degree vector, and $e$ is the total edge weight. The clustering objective is
\begin{equation}
    \mathcal{L}_{\text{cluster}} = -\frac{1}{2e} \sum_{k=1}^c \mathbf{c}_k^T \mathbf{B} \mathbf{c}_k 
    + \lambda \left\| \frac{1}{n} \mathbf{C}^T \mathbf{1} - \frac{1}{c} \mathbf{1} \right\|_2^2,
\end{equation}
where the first term encourages assignments that maximize modularity, and the second regularizes balanced use of the $c$ clusters. The modularity term pushes strongly interconnected nodes into the same cluster beyond random chance, while the regularization term prevents cluster collapse by encouraging balanced cluster utilization and reducing semantic redundancy.

Next, when clustering the nodes of $G^{(m)}$ to obtain $G'^{(m)}$, the cluster-level node features are:
\begin{equation}
    X_{\text{cluster}}^{(m, n+1)} = f_{\text{proj}, x} \left( \text{SELU} \left( \frac{(\mathbf{C}^{(m)})^T X^{(m, n)}}{N^{(m)}} \right) \right),
\end{equation}
where $m$ indexes the modality and we start from $X^{(m,0)}$ on $G^{(m)}$ ($n = 0$). The SELU transformation is controlled by a user-defined cluster hierarchy specifying the number of clusters per modality, with
$N^{(I)} < N^{(\mathbf{L})} < N^{(T)}$.
This reflects the assumed information density: the text graph uses the most clusters, the FC graph an intermediate number, and the image graph the fewest.

% \subsection{Graph Featurizer}

Given the clustered, modality-specific graph \(G'^{(m)} = (V'^{(m)}, E'^{(m)})\), we use a three-layer GCN featurization 
\(f_{\text{GCN}\times 3}(G'^{(m)})\),
to learn higher-order relational representations. Let \(X'^{(m)}_{0}\) be features of the clustered initial node $n$ for \(G'^{(m)}\), and define the normalized adjacency matrix with self-loops as $\widehat{\mathbf{A}} = \widetilde{\mathbf{D}}^{-0.5} \widetilde{\mathbf{A}} \widetilde{\mathbf{D}}^{-0.5}$. Each GCN layer \(l=0, 1, 2\) performs graph convolution, a linear transformation, batch normalization, and a ReLU activation:
\begin{equation}
{
\begin{aligned}
X'^{(m,l+1)}_{n+1}
&=
\operatorname{ReLU}\!\left(
\operatorname{BN}\!\left(
\widehat{\mathbf{A}} X^{(m,l)}_{n} \mathbf{W}^{(l)}
\right)\right),
\end{aligned}}
\label{eq:gcn_update}
\end{equation}

After graph featurization with GCN, we apply dropout and global average pooling during training to obtain modality-specific, graph-level embeddings \(g^{(m)}\). These embeddings are then projected onto a unit hypersphere latent space for contrastive learning, multimodal fusion, and downstream diagnostic classification.

\subsection{Stage 2: Contrastive Learning-Based Diagnosis}

% This stage performs diagnostic classification with multimodal embeddings. Modality-specific embeddings $g^{(m)}$ are projected into a shared latent space and refined into contrastive embeddings $z'^{(m)}$ via supervised contrastive learning. An attention-based fusion module integrates the most informative modalities, and the fused representation is passed to a classification head to predict the target class.

% \subsubsection{Component 4: Supervised Contrastive learning for the extracted multimodal features}
Given the modality-specific graph embeddings $g^{(m)}$ from the image, sparse FC, and text branches, we use supervised contrastive learning to jointly optimize their representations in a shared diagnostic latent space. For each subject $i$ and modality $m$, the embedding is passed through a modality-specific projection head and then $\ell_2$-normalized onto the unit hypersphere:
\[
z_i^{(m)} = \frac{\phi_{(m)}\bigl(g_i^{(m)}\bigr)}{\bigl\|\phi_{(m)}\bigl(g_i^{(m)}\bigr)\bigr\|_2},
\]
where $\phi_I(\cdot)$, $\phi_{\mathbf{L}}(\cdot)$, and $\phi_T(\cdot)$ are two-layer MLP projection heads for the image, sparse FC, and text modalities, respectively. We measure alignment between projected embeddings using cosine similarity $
s(z_i^a, z_j^b) = (z_i^a)^\top z_j^b,
$ where $a, b \in \{I, \mathbf{L}, T\}$ index the modality.

For a mini-batch of $B$ subjects, we use a directional InfoNCE loss to align projected representations from modalities $a$ and $b$. For each anchor in modality $a$, the embedding of the same subject in modality $b$ is the positive pair $(a, b^+)$, and embeddings in $b$ from other subjects are negative pairs $(a, b^-)$. The directional contrastive loss from $a$ to $b$ is
\begin{equation}
{
\mathcal{L}_{a \rightarrow b}
=
-\frac{1}{B}
\sum_{i=1}^{B}
\log
\left(
\frac{\mathcal{S}_{i}^{+}}
{\mathcal{Z}_{i}}
\right).
}
\label{eq:directional_contrastive_loss}
\end{equation}
Here, $\mathcal{S}_{i}^{+}=\exp\!\left(\mathbf{z}_{i}^{a}\cdot\mathbf{z}_{i}^{b^{+}}/\tau\right)$ is the exponentiated similarity between anchor $\mathbf{z}_{i}^{a}$ and its positive counterpart $\mathbf{z}_{i}^{b^{+}}$, and $\mathcal{Z}_{i}=\mathcal{S}_{i}^{+}+\sum_{b^{-}\in B}
\exp\!\left(\mathbf{z}_{i}^{a}\cdot\mathbf{z}_{i}^{b^{-}}/\tau\right)$ adds the positive-pair similarity to those with all negatives. The temperature $\tau$ controls the sharpness of the similarity distribution. To remove directional bias, we use the symmetric loss
\begin{equation}
    \mathcal{L}_{a,b} = \frac{1}{2} (\mathcal{L}_{a \to b} + \mathcal{L}_{b \to a}).
\end{equation}

The complete pairwise contrastive loss aligns all modality pairs as:
\begin{equation}
    \mathcal{L}_{\text{pair}} = \mathcal{L}_{I, \mathbf{L}} + \mathcal{L}_{I, T} + \mathcal{L}_{\mathbf{L}, T}
\end{equation}
Because the text modality contains interpretable phenotypic and brain-activation information, we further use it as a semantic anchor for the image and FC modalities and define the text-anchored loss as:
\begin{equation}
    \mathcal{L}_{\text{text}} = \frac{1}{2} (\mathcal{L}_{T, I} + \mathcal{L}_{T, \mathbf{L}})
\end{equation}
Thus, $\mathcal{L}_{\text{pair}}$ builds a shared multimodal embedding latent space, while the $\mathcal{L}_{\text{text}}$ organizes this space around the semantic structure provided by the text modality.

Using the diagnostic labels, we apply supervised contrastive learning to the text embeddings, encouraging subjects with the same diagnosis to cluster and those with different diagnoses to separate. We partition the multimodal latent space into two prototype regions based on each subject's ground-truth label, and define the supervised InfoNCE loss over the resulting structured text embedding space as:
\begin{equation}
{
\mathcal{L}_{\mathrm{supcon}}
=
-\sum_{i=1}^{B}
\frac{1}{|P(i)|}
\sum_{p \in P(i)}
\log
\left(
\frac{\mathcal{S}_{i,p}^{+}}
{\mathcal{Z}_{i}}
\right).
}
\label{eq:supcon_loss}
\end{equation}
where $\mathcal{S}_{i,p}^{+}=
\exp\!\left(
\mathbf{z}_{i}^{(T)}
\cdot
\mathbf{z}_{p}^{(T)}
/\tau
\right)$ is the exponentiated similarity between the anchor text embedding $\mathbf{z}_{i}^{(T)}$ and $\mathbf{z}_{p}^{(T)}$, the projected text embeddings of the disorder and control prototypes. The normalization term $\mathcal{Z}_{i}=\sum_{\substack{j=1,j \neq i}}^{B}\exp\!\left(\mathbf{z}_{i}^{(T)}\cdot\mathbf{z}_{j}^{(T)}/\tau \right)$ sums the exponentiated similarities between the anchor and all other batch samples, excluding itself. $\tau$ is the temperature parameter.

The final joint multimodal contrastive loss combining pairwise modality, text-anchored, and supervised diagnostic terms is:
\begin{equation}
    \mathcal{L}_{\text{con}} = \lambda_{\text{pair}}\mathcal{L}_{\text{pair}} + \lambda_{\text{text}}\mathcal{L}_{\text{text}} + \lambda_{\text{sup}}\mathcal{L}_{\text{supcon}},
\end{equation}
where $\lambda_{\text{pair}}$, $\lambda_{\text{text}}$, and $\lambda_{\text{sup}}$ weight each loss. Backpropagation with this joint loss updates the modality-specific projection heads, refining the original graph-based embeddings $g^{(m)}$, first projected as $z^{(m)}$, into improved modality-specific contrastive embeddings $z'^{(m)}$ for downstream multimodal fusion.

% \subsubsection{Component 5: Multi-Head Self-Attention Multimodal Fusion Module}

After contrastive learning, the aligned modality embeddings
$z'^{(m)}$ are element-wise fused into
$z_i^{s} = z_i'^{(I)} + z_i'^{(L)} + z_i'^{(T)}$,
preserving the shared embedding dimension $d_h$. This aligned embedding
is used as the query $Q_s$, while each modality embedding provides the
keys $K_m$ and values $V_m$ in computing attentions. This strategy allows
the weighting with the most task-relevant embedding, i.e., the aligned
embedding, rather than depending on any single modality embedding.
Attentions are computed as
\begin{equation}
\mathrm{attn}_{m \Leftrightarrow s}
=
\mathrm{softmax}
\left(
\frac{Q_s K_m^{\top}}{\sqrt{d_h}}
\right)
V_m .
\label{eq:modality_attention}
\end{equation}

From this, multiple parallel attentions with multiple heads are computed.
The $i$-th head for all modalities $m \in \{I,L,T\}$ is computed as
\[
\mathrm{head}_{m}^{i}
=
\mathrm{attn}_{m}
\left(
Q_s W_i^{Q_s},
K_m W_i^{K_m},
V_m W_i^{V_m}
\right),
\label{eq:attention_head}
\]
where $W_i^{Q_s}$, $W_i^{K_m}$, and $W_i^{V_m}$ are the projection weight
matrices for the $i$-th attention head. The modality-specific multi-head
self-attention (MMHSA) output is obtained by concatenating all heads as
\begin{equation}
\begin{aligned}
Y_m^{*}
&=
\mathrm{MultiHead}(Q_s,K_m,V_m) \\
&=
\mathrm{Concat}
\left(
\mathrm{head}_{m}^{1},\ldots,\mathrm{head}_{m}^{n}
\right)
W_m^{o}.
\end{aligned}
\label{eq:mmhsa_output}
\end{equation}
where $W_m^{o}$ is the output projection weight matrix, $n=10$ is the
number of attention heads, and $\mathrm{Concat}(\cdot)$ denotes head
concatenation. The resulting attention-refined modality representations are then passed
into a lightweight triple multimodal cyclic fusion network (TMCFN) \cite{zhang2022triple} for final multimodal fusion while avoiding excessive additional parameters
that may further increase the computational cost.

% \begin{figure}[tbp]
%     \centering
%     \includegraphics[width=\columnwidth]{Figure4_Attn_MultimodalFusion.pdf}
%     \caption{Attention-based Multimodal Fusion Module}
%     \label{Figure4_Attn_MultimodalFusion}
% \end{figure} 

% \subsubsection{Component 6: Diagnostic Classification Head}

The fused representation is passed to a classification head to produce the diagnostic probability distribution:
\begin{equation}
    \hat{y}_i = \text{softmax}(W_c f_i + b_c),
\end{equation}
where $\hat{y}_i$ denotes the predicted class probabilities (for binary diagnosis, $\hat{y}_i \in \mathbb{R}^2$). The classification loss is the cross-entropy:
\begin{equation}
    \mathcal{L}_{cls} = -\frac{1}{B} \sum_{i=1}^B \sum_{c=1}^C y_{i,c} \log (\hat{y}_{i,c}),
\end{equation}
with $C$ diagnostic classes. The final objective combines classification and contrastive losses:
\begin{equation}
    \mathcal{L}_{total} = \mathcal{L}_{cls} + \lambda_{con}\mathcal{L}_{con},
\end{equation}
where $\mathcal{L}_{cls}$ learns the diagnostic decision boundary and $\mathcal{L}_{con}$ aligns image, FC, and text modalities; $\lambda_{con}$ weights the contrastive term.

\section{Experimental Results}
% \subsection{Evaluation}

% Evaluation covers diagnostic performance, ablation effects, sensitivity, computational efficiency, and interpretability. Metrics include Accuracy (ACC), F1-score (F1), Sensitivity (SEN), Specificity (SPE), Area Under the ROC Curve (AUC-ROC), Davies-Bouldin Index (DBI), Calinski-Harabasz Index (CHI), and Silhouette Coefficient (SC).

% \subsection{Datasets}
We analyze five multisite rs-fMRI datasets spanning diverse brain disorders: ABIDE, ADHD-200, ADNI, REST-meta-MDD, and HCP-EP, each with rs-fMRI and phenotypic data. Preprocessing follows dataset-specific pipelines: CCS (ABIDE), Athena (ADHD-200), standard ADNI preprocessing, DPARSF \cite{yan2010dparsf} (REST-meta-MDD), and modified ABCD-BIDS \cite{feczko2021adolescent} (HCP-EP). Shared steps include motion correction, nuisance regression, MNI registration, temporal band-pass filtering, and spatial smoothing, with dataset-specific additions such as motion scrubbing (ADNI), initial-volume removal (REST-meta-MDD), and distortion/structural correction (HCP-EP). We used an 80/20 stratified train–test split balanced by site and diagnosis across datasets.

% We use all five datasets for diagnostic performance, ablation, and clinical interpretability. ABIDE is the main benchmark for sensitivity, efficiency, GGMM stability, cross-site generalization, missing-modality analysis, hyperparameter sensitivity, and latent-space visualization. SATD threshold sensitivity is evaluated on ABIDE, ADHD-200, and HCP-EP to compare with RTGMFF \cite{jia2025rtgmff} and to test generalization to a less commonly used dataset.

% \subsection{Implementation Details}
Experiments used Python 3.9.16, PyTorch 2.1, and PyTorch Geometric 2.4 on an NVIDIA RTX 4090 (CUDA 12.4). Image features came from frozen ViT-B/16 (196+1 [CLS] tokens, 768 dim) and text features from frozen BERT-base-uncased (768-dim tokens). T0-11B was fine-tuned for FVSerial. For LLSFC, $\alpha=0.01$, $\beta=0.1$, and modality graphs used $K=10$ nearest neighbors. With $(N^{(I)}<N^{(L)}<N^{(T)})$, modularity clustering used 32, 64, and 128 clusters for image, FC, and text. The 3-layer GCN had hidden sizes 64, 128 and 256, with dropout $p=0.3$, batch norm, and ReLU. Projection heads were 2-layer MLPs (256 hidden, 128 output). The contrastive temperature was $\tau=0.07$. Loss weights were $\lambda_{\mathrm{pair}}=1.0$, $\lambda_{\mathrm{text}}=0.5$, $\lambda_{\mathrm{sub}}=0.5$, and $\lambda_{\mathrm{con}}=0.4$. Training used AdamW (learning rate $10^{-4}$, weight decay $10^{-4}$), batch size 32, a cosine schedule, 200 epochs, and early stopping with patience 20. GGMM-based SATD used $n_{init}=5$ EM restarts, tolerance $\varepsilon=10^{-6}$, 200 maximum iterations, 2,000 grid points on $[0.001, 2.0]$, Brent–Dekker threshold refinement, and a minimum separation of 0.05 between $\tau_1$ and $\tau_2$.

%\FloatBarrier
\begin{table*}[!ht]
\caption{Diagnostic performance on five datasets, reported as mean$\pm$STD percentages over five runs. Best results are bolded. Modalities: I = \textit{f}MRI image, FC = functional connectivity, Txt = text, $\checkmark$ = used, $\times$ = not used.}
\label{table1_Diagnostic_performance}
\centering

\renewcommand{\arraystretch}{1.12}
\setlength{\tabcolsep}{2.8pt}

{\footnotesize
\resizebox{\textwidth}{!}{%
\begin{tabular}{|l|c|cc|cc|cc|cc|cc|}
\hline

\multirow{2}{*}{\textbf{Method}}
&
\multirow{2}{*}{\textbf{Modalities}}
&
\multicolumn{2}{c|}{\textbf{ABIDE}}
&
\multicolumn{2}{c|}{\textbf{ADHD-200}}
&
\multicolumn{2}{c|}{\textbf{ADNI}}
&
\multicolumn{2}{c|}{\textbf{HCP-EP}}
&
\multicolumn{2}{c|}{\textbf{REST-meta-MDD}}
\\
\cline{3-12}

&
&
\textbf{ACC} & \textbf{F1}
&
\textbf{ACC} & \textbf{F1}
&
\textbf{ACC} & \textbf{F1}
&
\textbf{ACC} & \textbf{F1}
&
\textbf{ACC} & \textbf{F1}
\\
\hline

\multicolumn{12}{|l|}{
\textbf{Unimodal: (i) Functional Connectivity (FC-Only) Baselines}
}
\\
\hline

SVM
& $\times~|~\checkmark~|~\times$
& $63.2\pm2.1$ & $62.8\pm2.3$
& $58.6\pm2.4$ & $57.9\pm2.6$
& $65.4\pm1.9$ & $64.7\pm2.0$
& $70.1\pm1.8$ & $69.5\pm1.9$
& $56.3\pm2.5$ & $55.7\pm2.7$
\\

GCN \cite{kipf2016semi}
& $\times~|~\checkmark~|~\times$
& $68.4\pm1.7$ & $67.9\pm1.8$
& $63.1\pm1.9$ & $62.4\pm2.1$
& $70.2\pm1.6$ & $69.6\pm1.7$
& $72.8\pm1.5$ & $74.3\pm1.6$
& $61.4\pm2.0$ & $60.8\pm2.2$
\\

GAT \cite{velivckovic2017graph}
& $\times~|~\checkmark~|~\times$
& $69.7\pm1.5$ & $69.2\pm1.7$
& $64.3\pm1.8$ & $63.7\pm1.9$
& $71.8\pm1.5$ & $71.2\pm1.6$
& $75.0\pm1.4$ & $75.5\pm1.5$
& $62.7\pm1.9$ & $62.1\pm2.0$
\\

GraphSAGE \cite{hamilton2017inductive}
& $\times~|~\checkmark~|~\times$
& $68.9\pm1.6$ & $68.4\pm1.7$
& $63.8\pm1.8$ & $63.2\pm2.0$
& $71.1\pm1.6$ & $70.5\pm1.7$
& $73.4\pm1.4$ & $74.9\pm1.5$
& $62.0\pm2.0$ & $61.4\pm2.1$
\\

BrainGB \cite{cui2022braingb}
& $\times~|~\checkmark~|~\times$
& $70.8\pm1.5$ & $70.3\pm1.6$
& $66.9\pm1.7$ & $66.3\pm1.8$
& $72.7\pm1.5$ & $72.1\pm1.6$
& $74.6\pm1.4$ & $77.1\pm1.4$
& $64.1\pm1.9$ & $63.5\pm2.0$
\\

Ada-FCN \cite{xun2025ada}
& $\times~|~\checkmark~|~\times$
& $71.6\pm1.4$ & $71.1\pm1.5$
& $67.4\pm1.6$ & $66.8\pm1.7$
& $73.6\pm1.4$ & $73.0\pm1.5$
& $73.5\pm1.3$ & $78.0\pm1.4$
& $64.9\pm1.8$ & $64.3\pm1.9$
\\

BrainGNN \cite{li2021braingnn}
& $\times~|~\checkmark~|~\times$
& $72.4\pm1.3$ & $71.9\pm1.4$
& $68.7\pm1.5$ & $68.1\pm1.6$
& $74.3\pm1.3$ & $73.7\pm1.4$
& $76.2\pm1.2$ & $78.7\pm1.3$
& $65.8\pm1.7$ & $65.2\pm1.8$
\\
\hline

\multicolumn{12}{|l|}{
\textbf{Unimodal: (ii) Functional Connectivity + Contrastive Learning
(FC+CL) Baselines}
}
\\
\hline

MGCL-DA \cite{zhang2025multi}
& $\times~|~\checkmark~|~\times$
& $75.6\pm1.1$ & $75.1\pm1.2$
& $71.2\pm1.3$ & $70.6\pm1.4$
& $77.3\pm1.2$ & $76.7\pm1.3$
& $79.1\pm1.1$ & $81.6\pm1.2$
& $68.7\pm1.5$ & $68.1\pm1.6$
\\

GCSC-TA \cite{zhang2025graph}
& $\times~|~\checkmark~|~\times$
& $74.8\pm1.2$ & $74.3\pm1.3$
& $70.4\pm1.4$ & $69.8\pm1.5$
& $76.5\pm1.2$ & $75.9\pm1.3$
& $76.3\pm1.1$ & $80.8\pm1.2$
& $67.9\pm1.6$ & $67.3\pm1.7$
\\

EC-GCL \cite{long2026intelligent}
& $\times~|~\checkmark~|~\times$
& $75.7\pm1.2$ & $75.2\pm1.3$
& $71.4\pm1.3$ & $70.8\pm1.4$
& $77.6\pm1.2$ & $77.0\pm1.3$
& $77.4\pm1.0$ & $81.9\pm1.1$
& $68.8\pm1.5$ & $68.2\pm1.6$
\\
\hline

\multicolumn{12}{|l|}{
\textbf{Unimodal: (iii) Functional Connectivity + Transformer
(FC+Transformer) Baselines}
}
\\
\hline

ContrasFormer \cite{xu2024contrasformer}
& $\times~|~\checkmark~|~\times$
& $73.3\pm1.1$ & $72.8\pm1.2$
& $70.9\pm1.3$ & $69.3\pm1.4$
& $70.1\pm1.1$ & $69.5\pm1.2$
& $77.2\pm1.0$ & $76.5\pm1.1$
& $69.4\pm1.5$ & $68.8\pm1.1$
\\

FBNetGen \cite{kan2022fbnetgen}
& $\times~|~\checkmark~|~\times$
& $73.8\pm1.5$ & $73.1\pm1.7$
& $71.3\pm1.2$ & $70.7\pm1.3$
& $72.2\pm1.1$ & $71.0\pm1.3$
& $78.5\pm1.4$ & $78.0\pm1.1$
& $69.9\pm1.4$ & $69.1\pm1.5$
\\

Com-BrainTF \cite{bannadabhavi2023community}
& $\times~|~\checkmark~|~\times$
& $74.1\pm1.1$ & $73.6\pm1.2$
& $72.1\pm1.3$ & $71.0\pm1.4$
& $72.9\pm1.8$ & $71.3\pm1.5$
& $79.1\pm1.1$ & $78.3\pm1.1$
& $70.2\pm1.1$ & $69.5\pm1.4$
\\

BrainNetTF \cite{kan2022brain}
& $\times~|~\checkmark~|~\times$
& $75.4\pm1.0$ & $74.9\pm1.1$
& $72.8\pm1.2$ & $72.0\pm1.6$
& $74.2\pm1.3$ & $73.6\pm1.1$
& $79.8\pm1.6$ & $78.6\pm1.1$
& $71.5\pm1.7$ & $69.9\pm1.8$
\\
\hline

\multicolumn{12}{|l|}{\textbf{Unimodal: Image}}
\\
\hline

CNN
& $\checkmark~|~\times~|~\times$
& $65.3\pm1.8$ & $64.7\pm1.9$
& $61.2\pm2.1$ & $60.6\pm2.2$
& $67.8\pm1.7$ & $67.2\pm1.8$
& $68.6\pm1.6$ & $67.0\pm1.7$
& $58.4\pm2.2$ & $57.8\pm2.3$
\\

MobileNetV2
& $\checkmark~|~\times~|~\times$
& $66.8\pm1.7$ & $66.2\pm1.8$
& $62.4\pm2.0$ & $61.8\pm2.1$
& $69.1\pm1.6$ & $68.5\pm1.7$
& $67.0\pm1.5$ & $66.4\pm1.6$
& $59.7\pm2.1$ & $59.1\pm2.2$
\\

VGG16
& $\checkmark~|~\times~|~\times$
& $65.9\pm1.8$ & $65.3\pm1.9$
& $61.7\pm2.0$ & $60.8\pm2.1$
& $68.4\pm1.7$ & $67.8\pm1.8$
& $69.3\pm1.5$ & $68.7\pm1.6$
& $58.9\pm2.2$ & $58.3\pm2.3$
\\

ViT-Image
& $\checkmark~|~\times~|~\times$
& $67.2\pm1.7$ & $66.6\pm1.8$
& $63.4\pm1.9$ & $62.8\pm2.0$
& $69.7\pm1.6$ & $69.1\pm1.7$
& $72.3\pm1.5$ & $72.7\pm1.6$
& $60.5\pm2.1$ & $59.9\pm2.2$
\\

BrainNetCNN \cite{kawahara2017brainnetcnn}
& $\checkmark~|~\times~|~\times$
& $70.1\pm1.6$ & $69.5\pm1.7$
& $66.2\pm1.8$ & $65.6\pm1.9$
& $72.4\pm1.5$ & $71.8\pm1.6$
& $70.0\pm1.4$ & $69.4\pm1.5$
& $63.2\pm1.9$ & $62.6\pm2.0$
\\
\hline

\multicolumn{12}{|l|}{\textbf{Unimodal: Text}}
\\
\hline

BERT-Text
& $\times~|~\times~|~\checkmark$
& $74.6\pm1.2$ & $74.1\pm1.3$
& $71.3\pm1.4$ & $70.7\pm1.5$
& $75.8\pm1.2$ & $74.2\pm1.3$
& $76.9\pm1.1$ & $75.4\pm1.2$
& $68.2\pm1.5$ & $67.6\pm1.6$
\\
\hline

\multicolumn{12}{|l|}{
\textbf{Multimodal: (i) Late Fusion Baselines}
}
\\
\hline

Ensemble$_1$
& $\checkmark~|~\checkmark~|~\times$
& $73.5\pm1.4$ & $73.0\pm1.5$
& $69.8\pm1.6$ & $69.2\pm1.7$
& $75.8\pm1.3$ & $75.2\pm1.4$
& $77.6\pm1.2$ & $76.1\pm1.3$
& $67.1\pm1.7$ & $66.5\pm1.8$
\\

Ensemble$_2$
& $\checkmark~|~\checkmark~|~\checkmark$
& $76.2\pm1.1$ & $75.7\pm1.2$
& $72.1\pm1.3$ & $71.5\pm1.4$
& $78.4\pm1.1$ & $77.8\pm1.2$
& $78.3\pm1.0$ & $77.8\pm1.1$
& $69.5\pm1.5$ & $68.9\pm1.6$
\\
\hline

\multicolumn{12}{|l|}{
\textbf{Multimodal: (ii) Intermediate Fusion Baselines}
}
\\
\hline

CMT
& $\checkmark~|~\checkmark~|~\checkmark$
& $78.4\pm1.0$ & $77.9\pm1.1$
& $74.8\pm1.2$ & $73.2\pm1.3$
& $76.6\pm1.0$ & $75.0\pm0.9$
& $78.5\pm0.9$ & $77.0\pm1.0$
& $71.7\pm1.4$ & $70.1\pm1.5$
\\

Concat-MLP
& $\checkmark~|~\checkmark~|~\checkmark$
& $76.8\pm1.2$ & $75.3\pm1.3$
& $72.6\pm1.3$ & $71.0\pm1.4$
& $78.0\pm1.2$ & $77.4\pm1.3$
& $80.9\pm1.1$ & $79.4\pm1.2$
& $70.0\pm1.5$ & $69.4\pm1.6$
\\

TUMSyn \cite{wang2025toward}
& $\checkmark~|~\times~|~\checkmark$
& $79.8\pm0.9$ & $78.3\pm1.0$
& $75.9\pm1.1$ & $74.3\pm1.2$
& $77.9\pm1.0$ & $76.3\pm1.1$
& $79.8\pm0.9$ & $78.3\pm1.0$
& $74.0\pm1.3$ & $73.4\pm1.4$
\\

BrainPrompt+ \cite{xu2026brainprompt+}
& $\times~|~\checkmark~|~\checkmark$
& $79.1\pm1.0$ & $78.6\pm1.1$
& $75.4\pm1.2$ & $74.8\pm1.3$
& $79.3\pm1.0$ & $78.7\pm1.2$
& $82.2\pm0.9$ & $81.7\pm1.0$
& $72.3\pm1.3$ & $71.7\pm1.4$
\\

PMIL \cite{choi2025pmil}
& $\times~|~\checkmark~|~\checkmark$
& $82.2\pm0.9$ & $81.7\pm1.0$
& $76.3\pm1.1$ & $75.7\pm1.2$
& $79.4\pm0.9$ & $79.8\pm1.7$
& $81.1\pm0.8$ & $80.6\pm0.9$
& $73.4\pm1.3$ & $72.8\pm1.4$
\\

fTSPL \cite{wang2024ftspl}
& $\checkmark~|~\checkmark~|~\checkmark$
& $81.7\pm0.9$ & $80.2\pm1.0$
& $76.8\pm1.1$ & $75.2\pm1.2$
& $80.9\pm0.9$ & $79.3\pm1.0$
& $82.6\pm0.8$ & $81.1\pm0.9$
& $74.9\pm1.2$ & $73.3\pm1.3$
\\

RTGMFF \cite{jia2025rtgmff}
& $\checkmark~|~\checkmark~|~\checkmark$
& $82.4\pm0.9$ & $81.9\pm1.0$
& $77.9\pm1.0$ & $76.3\pm1.1$
& $81.7\pm0.9$ & $80.0\pm0.6$
& $84.4\pm0.8$ & $83.9\pm0.9$
& $75.6\pm1.2$ & $74.0\pm1.3$
\\
\hline

\textbf{fMRI-TAMCL}
& $\checkmark~|~\checkmark~|~\checkmark$
& \textbf{84.7$\pm$0.8}
& \textbf{83.2$\pm$0.7}
& \textbf{80.1$\pm$1.3}
& \textbf{79.6$\pm$1.0}
& \textbf{83.3$\pm$0.7}
& \textbf{82.8$\pm$0.8}
& \textbf{86.4$\pm$0.7}
& \textbf{85.9$\pm$0.8}
& \textbf{78.6$\pm$1.1}
& \textbf{77.0$\pm$1.2}
\\
\hline

\end{tabular}%
}
}

\end{table*}
%\FloatBarrier

% \subsection{Baselines}

We compare fMRI-TAMCL with 20 unimodal and 9 multimodal baseline methods. SATD module was calibrated at subject level for fMRI-TAMCL. The yielded threshold was $\tau{_1}=[0.128\pm0.047]$ and 
$\tau{_2}=[0.263\pm0.061]$. Unimodal baselines include FC methods using Pearson correlation matrices, image-based CNN and ViT models, and a BERT text model trained on SATD+FVSerial. Multimodal baselines include late- and intermediate-fusion ensembles, CMT, Concat-MLP, TUMSyn, BrainPrompt+, PMIL, fTSPL \cite{wang2024ftspl}, and RTGMFF \cite{jia2025rtgmff}. RTGMFF is the closest baseline, as it also performs ROI-driven fMRI text generation but uses globally fixed thresholds across subjects. 

Table \ref{table1_Diagnostic_performance} shows that fMRI-TAMCL achieves the best diagnostic performance on all five datasets, with accuracies of 84.7\%, 80.1\%, 83.3\%, 86.4\%, and 78.6\% on ABIDE, ADHD-200, ADNI, HCP-EP, and REST-meta-MDD, respectively, surpassing RTGMFF \cite{jia2025rtgmff} by 1.6–3.0\%. The largest gain on HCP-EP suggests strong generalization across heterogeneous brain disorders.

\begin{figure}[!hb]
\centering

\begin{subfigure}[t]{0.46\columnwidth}
    \centering
    \includegraphics[width=\linewidth]{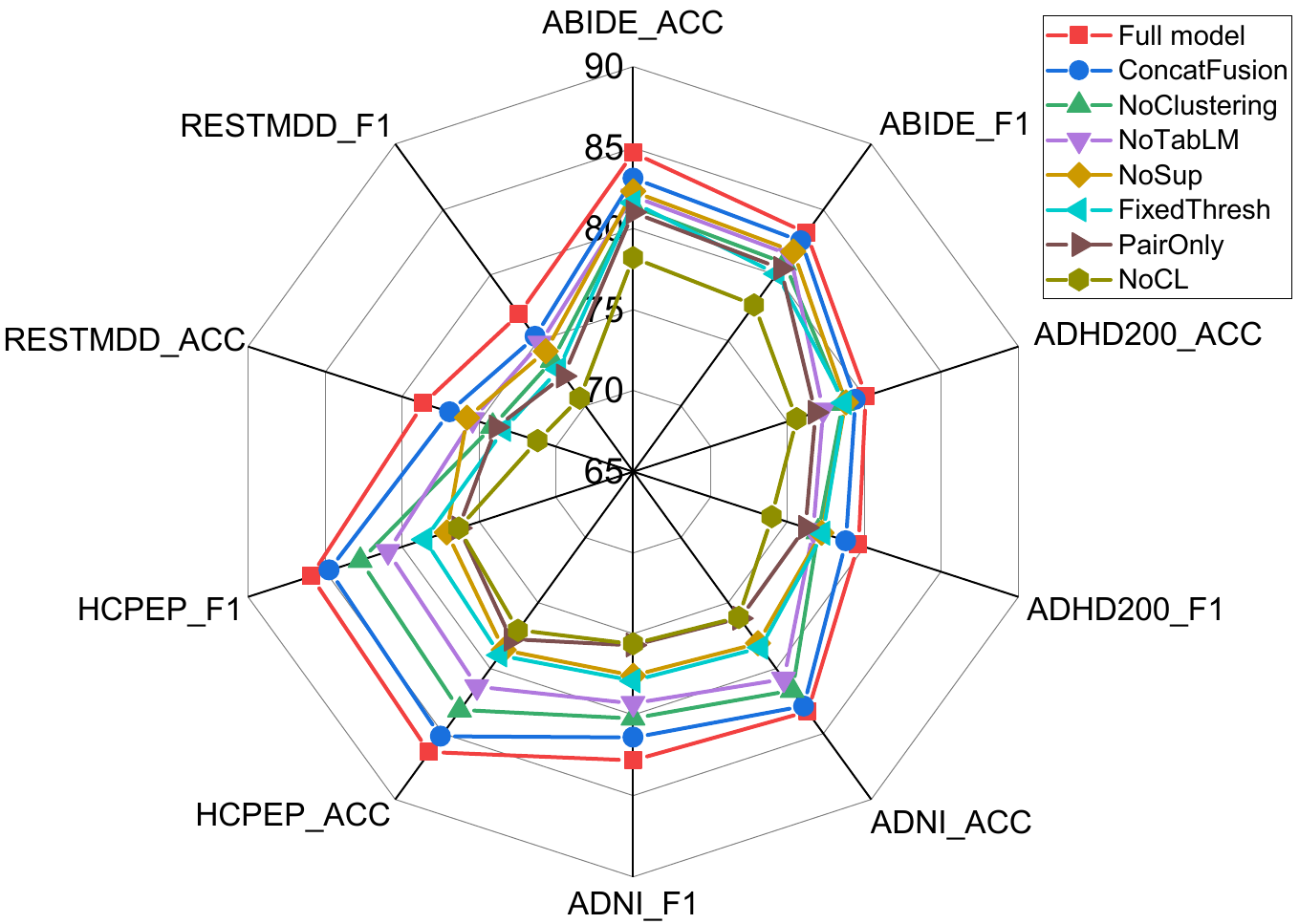}
    \caption{ACC and F1}
    \label{Figure_ablation_ACC_F1_radar}
\end{subfigure}
\hfill
\begin{subfigure}[t]{0.46\columnwidth}
    \centering
    \includegraphics[width=\linewidth]{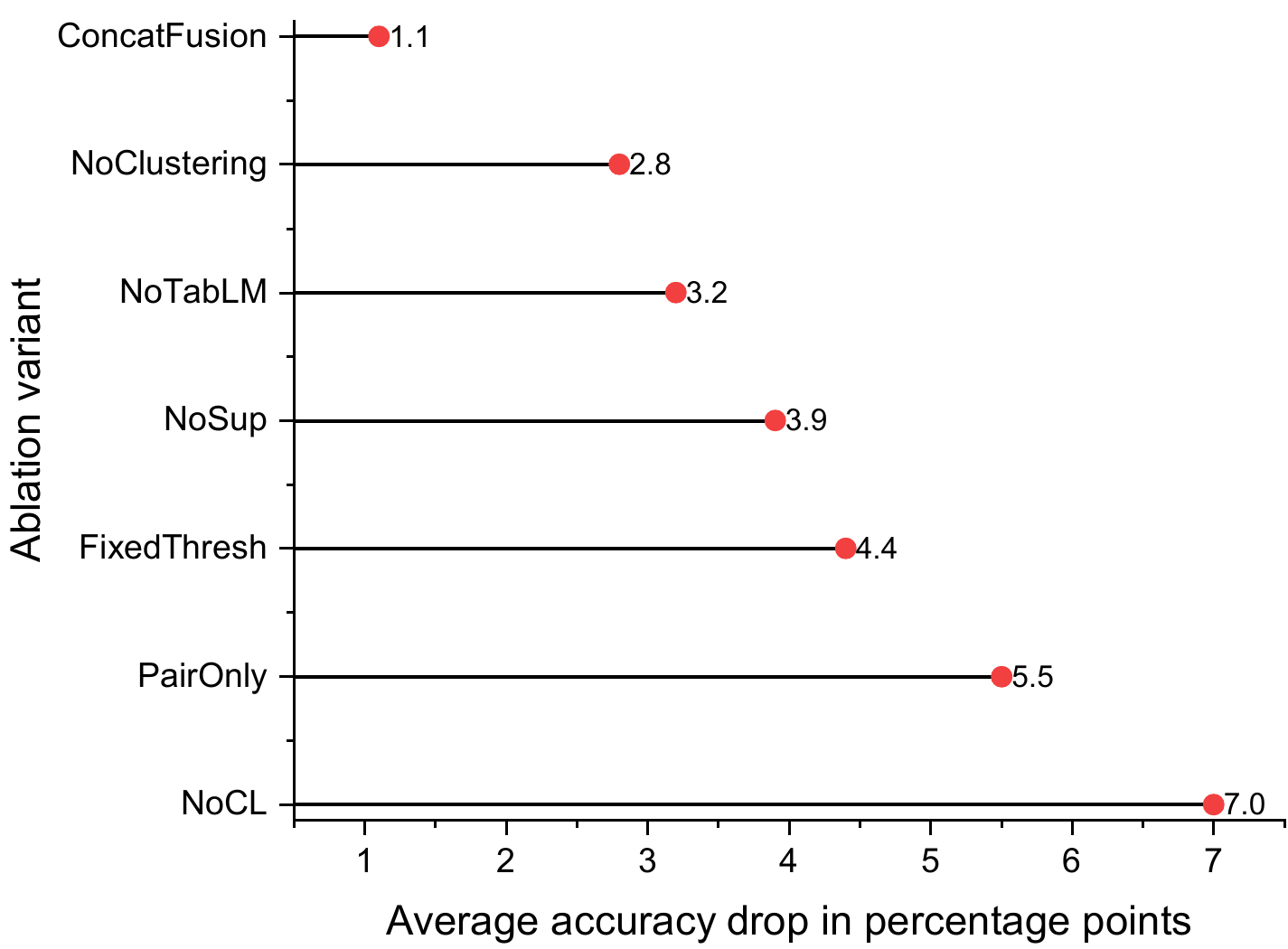}
    \caption{Average drop}
    \label{Figure5_ablation_avg_drop}
\end{subfigure}

\caption{Ablation study on five datasets, where each variant removes one architectural component from the full fMRI-TAMCL model.}
\label{Figure5_ablation_study}

\end{figure}

We conducted an ablation study comparing the full model with seven variants that remove or replace contrastive learning, text anchoring, supervised contrastive loss, SATD, LLM-based phenotypic serialization, modularity-based clustering, and attention-based fusion. As shown in Figure \ref{Figure5_ablation_study}, the full model achieves the best accuracy on all datasets. Contrastive learning is most critical: removing it causes the largest average drop (7\%), especially on HCP-EP. Text anchoring and supervised contrastive learning are also important, with average drops of 5.5\% and 3.9\%. Removing LLM-based phenotypic serialization, graph clustering, and attention fusion causes smaller average drops of 3.2\%, 2.8\%, and 1.1\%, indicating these components provide gains.

% \subsection{Cross-Site Generalization}
Leave-One-Site-Out cross-validation (LOSOCV) on ABIDE evaluates cross-site generalization by training fMRI-TAMCL on 16 sites and testing on the remaining one across all 17 sites. As shown in Table \ref{table3_LOSOCV}, fMRI-TAMCL attains the best LOSOCV performance (76.4\% ACC, 77.1\% SEN, 78.7\% SPE, 0.851 AUC-ROC, 76.2\% F1), surpassing RTGMFF \cite{jia2025rtgmff} by 2.2\% ACC, 3.2\% SEN, 2.9\% SPE, 0.032 AUC-ROC, and 1.6\% F1.

\begin{table}[!ht]
\caption{LOSOCV results on ABIDE. Mean$\pm$STD across
17 held-out sites. Best results are in bold.}
\label{table3_LOSOCV}
\centering
\renewcommand{\arraystretch}{1.10}
\setlength{\tabcolsep}{2.5pt}
\resizebox{\columnwidth}{!}{%
\begin{tabular}{|l|c|c|c|c|c|}
\hline
\textbf{Method} &
\textbf{ACC (\%)} &
\textbf{SEN (\%)} &
\textbf{SPE (\%)} &
\textbf{AUC-ROC} &
\textbf{F1 (\%)} \\
\hline

\multicolumn{6}{|l|}{\textbf{Unimodal}} \\
\hline
BrainNetCNN \cite{kawahara2017brainnetcnn}
& 62.3$\pm$4.2
& 60.8$\pm$5.1
& 63.8$\pm$4.7
& 0.668$\pm$0.045
& 62.0$\pm$4.3 \\

BrainGNN \cite{li2021braingnn}
& 64.7$\pm$3.9
& 63.2$\pm$4.6
& 66.3$\pm$4.3
& 0.693$\pm$0.042
& 64.5$\pm$3.9 \\

BERT-Text
& 66.4$\pm$2.9
& 64.9$\pm$3.8
& 67.1$\pm$3.7
& 0.718$\pm$0.031
& 65.7$\pm$3.1 \\

BrainNetTF
& 67.9$\pm$2.9
& 66.8$\pm$2.2
& 69.1$\pm$2.8
& 0.731$\pm$0.021
& 67.9$\pm$2.3 \\

EC-GCL
& 68.2$\pm$3.6
& 67.0$\pm$4.2
& 69.5$\pm$3.9
& 0.736$\pm$0.038
& 68.0$\pm$3.6 \\
\hline

\multicolumn{6}{|l|}{\textbf{Multimodal}} \\
\hline
TUMSyn
& 70.3$\pm$2.5
& 69.1$\pm$3.3
& 71.2$\pm$3.2
& 0.762$\pm$0.034
& 69.1$\pm$3.2 \\

BrainPrompt+
& 69.8$\pm$2.8
& 68.3$\pm$3.4
& 70.6$\pm$4.1
& 0.741$\pm$0.022
& 68.5$\pm$2.5 \\

PMIL
& 73.9$\pm$3.1
& 72.9$\pm$2.2
& 74.8$\pm$3.1
& 0.808$\pm$0.023
& 72.6$\pm$2.8 \\

fTSPL  \cite{wang2024ftspl}
& 72.4$\pm$3.2
& 71.2$\pm$3.6
& 73.6$\pm$3.8
& 0.787$\pm$0.041
& 71.4$\pm$3.7 \\

RTGMFF \cite{jia2025rtgmff}
& 74.2$\pm$3.3
& 73.9$\pm$3.9
& 75.8$\pm$3.5
& 0.819$\pm$0.034
& 74.6$\pm$3.4 \\

\textbf{fMRI-TAMCL}
& \textbf{76.4$\pm$2.1}
& \textbf{77.1$\pm$2.3}
& \textbf{78.7$\pm$2.1}
& \textbf{0.851$\pm$0.028}
& \textbf{76.2$\pm$2.2} \\
\hline
\end{tabular}%
}
\end{table}

% \subsection{Sensitivity Analysis}
% Sensitivity analysis evaluates fMRI-TAMCL under missing modalities, SATD threshold variations, and key hyperparameter changes, including contrastive loss weights and graph cluster numbers.
% \subsubsection{Missing Modality Sensitivity}
Table \ref{tabel4_Missing-modality} shows the robustness to missing modalities by removing image, FC, or text inputs, mirroring clinical scenarios such as motion-degraded FC. Unavailable baselines are marked N/A, and a phenotypic/demographic-text-only setting evaluates reliance on clinical text. fMRI-TAMCL achieves 79.5\% accuracy without images, 77.8\% without FC, 78.2\% without full text, and 81.8\% with only phenotypic/demographic text, outperforming RTGMFF \cite{jia2025rtgmff} by 2.3–5.5\% under comparable missing-modality conditions. It attains the highest stability-adjusted accuracy of 83.9\% with an average drop of only 4.6\%. Similar performance without FC and without text suggests both modalities contribute complementary, equally important diagnostic information.

\begin{table}[!tb]
\caption{Missing-modality sensitivity on ABIDE. Avg Drop from $ACC_{\mathrm{stable}}=\bar{x}_{\mathrm{ACC}}-\sigma_{\mathrm{ACC}}$. Best results are in bold.}
\label{tabel4_Missing-modality}
\centering

\renewcommand{\arraystretch}{1.15}
\setlength{\tabcolsep}{2.0pt}

{\footnotesize
\resizebox{\columnwidth}{!}{%
\begin{tabular}{|l|c|c|c|c|c|c|}
\hline

\textbf{Method}
&
\textbf{\shortstack{Miss\\Image\\ACC (\%)}}
&
\textbf{\shortstack{Miss\\FC\\ACC (\%)}}
&
\textbf{\shortstack{Miss Complete\\Text\\ACC (\%)}}
&
\textbf{\shortstack{Only Phenotypic/\\Demographic Text\\ACC (\%)}}
&
\textbf{\shortstack{All\\Modalities\\ACC (\%)}}
&
\textbf{\shortstack{Avg\\Drop}}
\\
\hline

TUMSyn
& 71.56
& N/A
& N/A
& 74.6
& 78.9
& 5.8
\\

BrainPrompt+
& N/A
& N/A
& N/A
& 73.8
& 78.1
& 4.3
\\

PMIL
& N/A
& 74.4
& 74.7
& N/A
& 81.3
& 6.8
\\

fTSPL \cite{wang2024ftspl}
& 75.6
& 73.1
& 73.9
& N/A
& 80.8
& 6.6
\\

RTGMFF \cite{jia2025rtgmff}
& 77.2
& N/A
& 75.7
& 76.3
& 81.5
& 5.1
\\
\hline

\textbf{fMRI-TAMCL}
& \textbf{79.5}
& \textbf{77.8}
& \textbf{78.2}
& \textbf{81.8}
& \textbf{83.9}
& \textbf{4.6}
\\
\hline

\end{tabular}%
}
}

\end{table}

% \subsubsection{SATD Threshold Sensitivity}

% \subsubsection{Hyperparameter Sensitivity}
% \noindent\textbf{\textit{i. Contrastive Loss Weight Sensitivity}}

We studied two hyperparameters: (i) contrastive loss weights and (ii) graph cluster count. On ABIDE (Table \ref{table4_SATD_sensitivity}), contrastive loss experiments show stable accuracy (82.8\%-84.7\%) for $\lambda_{con}=0.4$ across $\lambda_{pair}$, $\lambda_{text}$, and $\lambda_{sup}$. All settings outperform RTGMFF \cite{jia2025rtgmff}, with the best configuration $(\lambda_{pair}=1.0$, $\lambda_{text}=\lambda_{sup}=0.5)$, reaching 84.7\%. This suggests that fMRI-TAMCL is robust to loss-weight variation and benefits from both its architecture and contrastive objective. Results on graph cluster counts show the best are 32, 64, and 128 clusters for image, FC, and text, respectively. Increasing or decreasing these counts reduces accuracy. These results highlight the importance of the modularity-based graph clustering module in fMRI-TAMCL.

\begin{table}[!ht]
\caption{Hyperparameter sensitivity on ABIDE. Bolded rows are optimal settings used in all experiments.}
\label{table4_SATD_sensitivity}
\centering

\renewcommand{\arraystretch}{1.15}
\setlength{\tabcolsep}{16pt}

\begin{tabular}{|l|c|c|}
\hline
\textbf{Hyperparameter}
&
\textbf{ACC (\%)}
&
\textbf{F1}
\\
\hline

\multicolumn{3}{|l|}{
\textbf{Contrastive loss weight sensitivity
$(\lambda_{\mathrm{pair}}, \lambda_{\mathrm{text}}, \lambda_{\mathrm{sup}})$}
}
\\
\hline

\quad $(0.5, 0.5, 0.5)$
& $83.1\pm0.9$
& $83.0\pm1.0$
\\

\quad \textbf{$(1.0, 0.5, 0.5)$}
& \textbf{$84.7\pm0.8$}
& \textbf{$83.2\pm0.7$}
\\

\quad $(0.5, 0.5, 1.0)$
& $83.9\pm0.9$
& $83.8\pm0.9$
\\
\hline

\multicolumn{3}{|l|}{
\textbf{Number of Graph Clusters
$(N^{(I)}, N^{(L)}, N^{(T)})$}
}
\\
\hline

\quad \textbf{$(32, 64, 128)$}
& \textbf{$84.7\pm0.8$}
& \textbf{$83.2\pm0.7$}
\\

\quad $(128, 64, 32)$
& $79.3\pm1.1$
& $79.2\pm1.2$
\\
\hline

\end{tabular}

\end{table}

% \subsection{GGMM Convergence and Fallback Analysis}
Table \ref{table7_GGMM_convergence} shows the GGMM-based SATD performance across all ABIDE subjects. The module fully converges for 95.4\% of subjects and fully or partially handles 98.5\%, with only 1.5\% requiring fallback thresholds. Excluding fallback cases raises accuracy from 84.7\% to 85.6\%, indicating that fallback subjects add slight thresholding noise but are too rare to compromise SATD’s overall reliability and robustness.

\begin{table}[!tb]
\caption{GGMM convergence across all ABIDE subjects ($N=1{,}112$).}
\label{table7_GGMM_convergence}
\centering

\renewcommand{\arraystretch}{1.15}
\setlength{\tabcolsep}{5pt}

{\footnotesize
\begin{tabular}{
|p{0.48\columnwidth}
|p{0.16\columnwidth}
|p{0.22\columnwidth}|}
\hline

\textbf{SATD Outcome}
&
\textbf{\shortstack[l]{Count ($N$)
}}
&
\textbf{\shortstack[l]{Proportion (\%)
}}

\\
\hline

Full convergence (2 crossing points)
&
1,061
&
95.4
\\
\hline

Partial (1 crossing point; $\tau_2$ from mean scaling)
&
34
&
3.1
\\
\hline

Fallback (percentile-based thresholds)
&
17
&
1.5
\\
\hline

\end{tabular}
}
\end{table}

% \subsection{Computational Efficiency Analysis}
Table \ref{table8_computational_Complexity} reports the computational cost and conservatively stability-adjusted accuracy of fMRI-TAMCL and key multimodal baselines on ABIDE, including trainable parameters, training time per epoch, and inference time per subject. fMRI-TAMCL attains the highest accuracy with a moderate footprint: 21.2M parameters, 81 seconds of training per epoch, and 22.6 ms of inference per subject, improving accuracy by 3.3\% absolute over RTGMFF \cite{jia2025rtgmff} with only modest increases in model size and latency.  
% Future work will validate fMRI-TAMCL on real-world clinical cohorts by training on public multisite datasets and testing on hospital datasets as fully held-out sites. Preliminary analysis suggests that the main computational overhead arises from three-modality graph construction and clustering, which can be precomputed and cached to substantially reduce deployment-time inference cost.

\begin{table}[!tb]
\caption{Comparison of computational efficiency (training and inference times) and conservative stability-adjusted accuracy
$ACC_{\mathrm{stable}}$.}
\label{table8_computational_Complexity}
\centering

\renewcommand{\arraystretch}{1}
\setlength{\tabcolsep}{2.0pt}

{\scriptsize
\resizebox{\columnwidth}{!}{%
\begin{tabular}{|l|c|c|c|c|}
\hline

\textbf{Multimodal Method}
&
\textbf{\shortstack{Parameters\\(M)}}
&
\textbf{\shortstack{Train Time/\\Epoch (s)}}
&
\textbf{\shortstack{Inference Time/\\Subject (ms)}}
&
\textbf{\shortstack{$ACC_{\mathrm{stable}}$\\(\%)}}
\\
\hline

TUMSyn
& 15.1
& 78
& 15.1
& 78.9
\\

BrainPrompt+
& 17.8
& 62
& 13.2
& 78.1
\\

PMIL
& 18.3
& 67
& 21.3
& 81.3
\\

fTSPL \cite{wang2024ftspl}
& 18.7
& 74
& 19.9
& 80.8
\\

RTGMFF \cite{jia2025rtgmff}
& 20.7
& 70
& 17.9
& 81.5
\\
\hline

\textbf{fMRI-TAMCL}
& \textbf{21.2}
& \textbf{81}
& \textbf{22.6}
& \textbf{83.9}
\\
\hline

\end{tabular}%
}
}

\end{table}

\begin{figure*}[!t]
\centering

\begin{subfigure}[t]{0.19\textwidth}
    \centering
    \includegraphics[width=\linewidth]{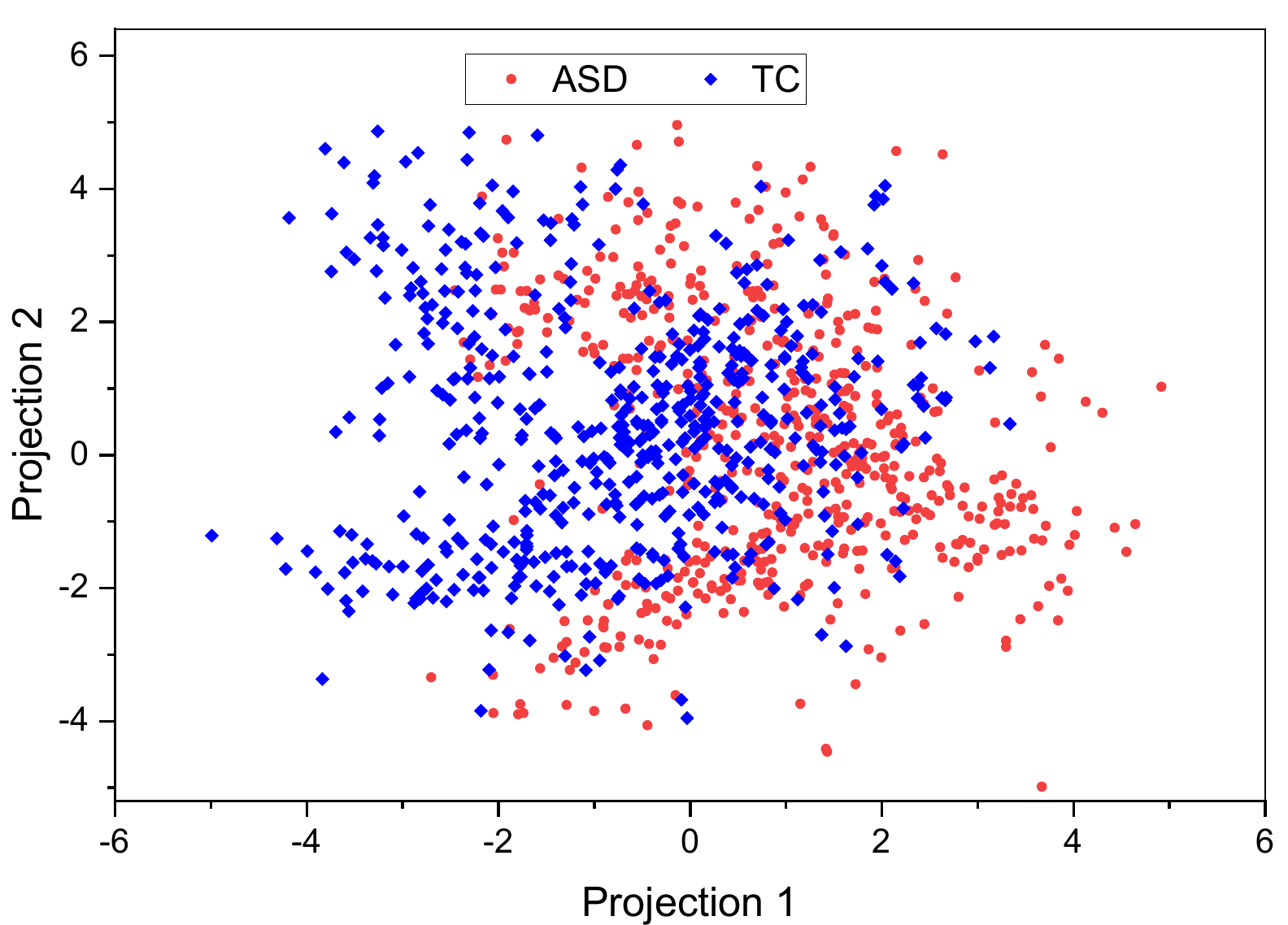}
    \caption{TAMCL-NoCL \\
    {\tiny DBI=1.952, CHI=108.4, SC=0.112}}
    \label{Figure6a_umap_no_cl}
\end{subfigure}
\hfill
\begin{subfigure}[t]{0.19\textwidth}
    \centering
    \includegraphics[width=\linewidth]{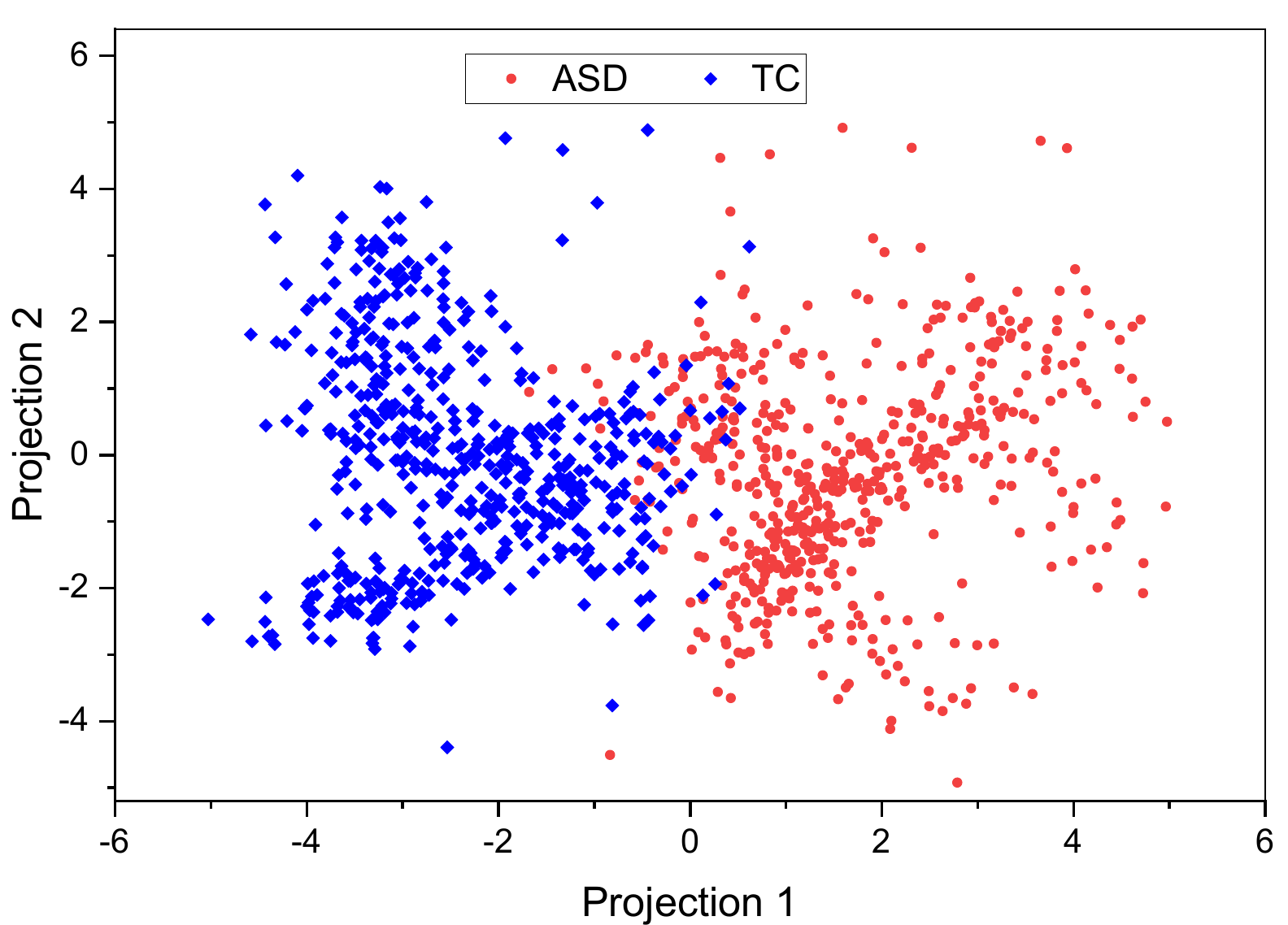}
    \caption{TAMCL-PairOnly \\
    {\tiny DBI=1.611, CHI=145.6, SC=0.185}}        \label{Figure6b_umap_pair_only}
\end{subfigure}
\hfill
\begin{subfigure}[t]{0.19\textwidth}
    \centering
    \includegraphics[width=\linewidth]{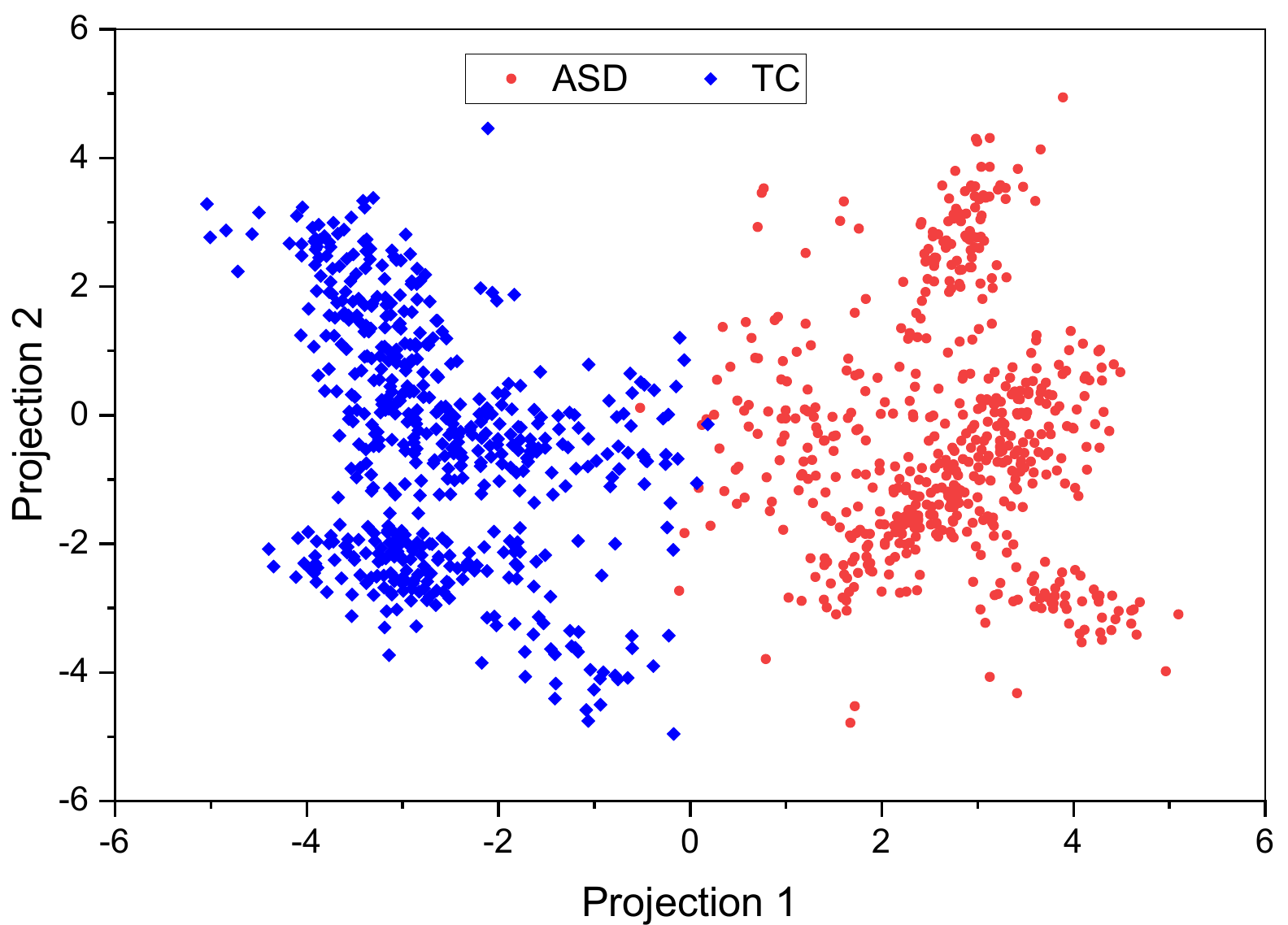}
    \caption{TAMCL-NoSup \\
    {\tiny DBI=1.595, CHI=180.2, SC=0.234}}
    \label{Figure6c_umap_no_sup}
\end{subfigure}
\hfill
\begin{subfigure}[t]{0.19\textwidth}
    \centering
    \includegraphics[width=\linewidth]{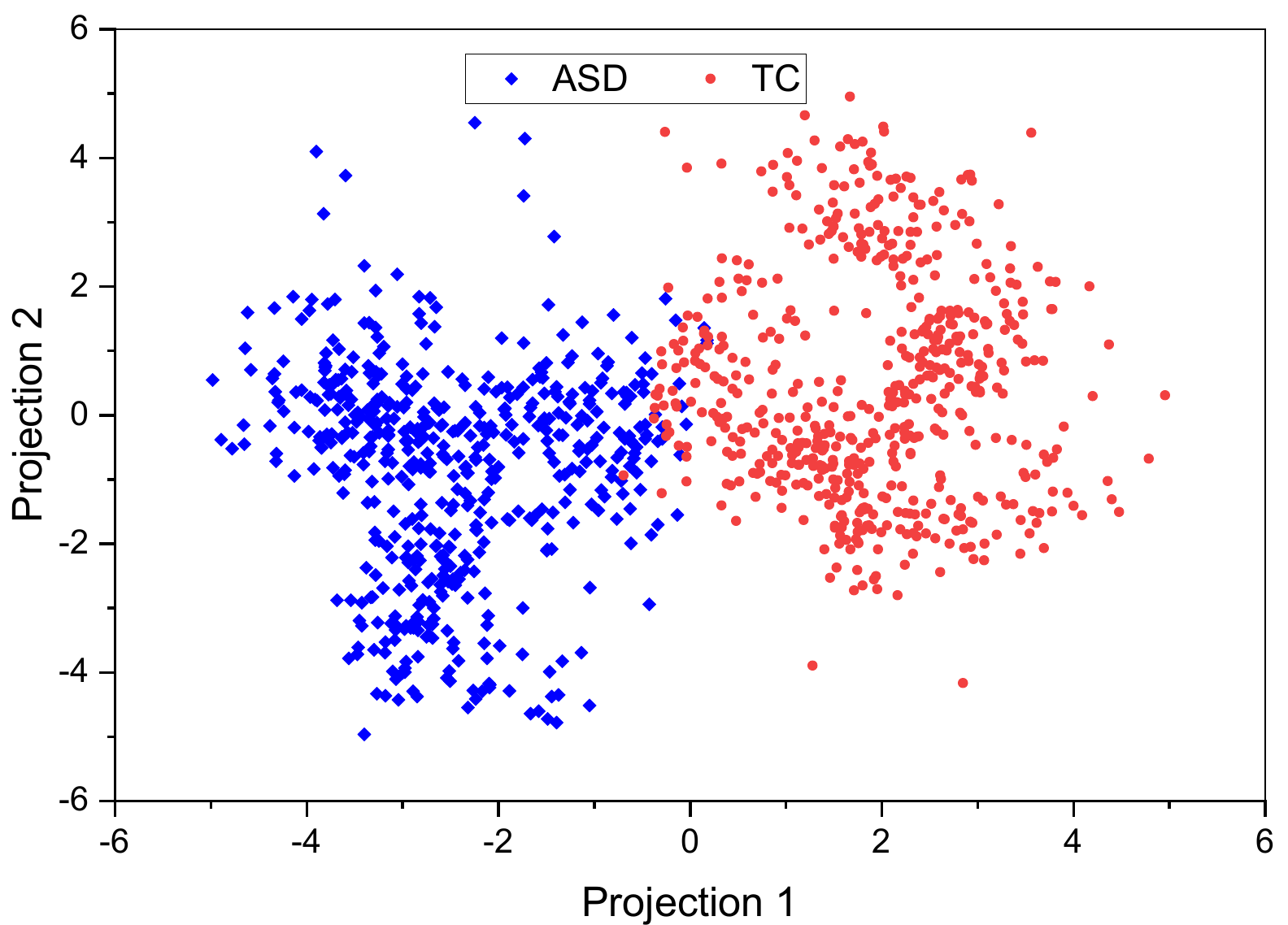}
    \caption{RTGMFF \cite{jia2025rtgmff} \\
    {\tiny DBI=1.387, CHI=214.8, SC=0.341}
    }
    \label{Figure6d_umap_rtgmff}
\end{subfigure}
\hfill
\begin{subfigure}[t]{0.19\textwidth}
    \centering
    \includegraphics[width=\linewidth]{    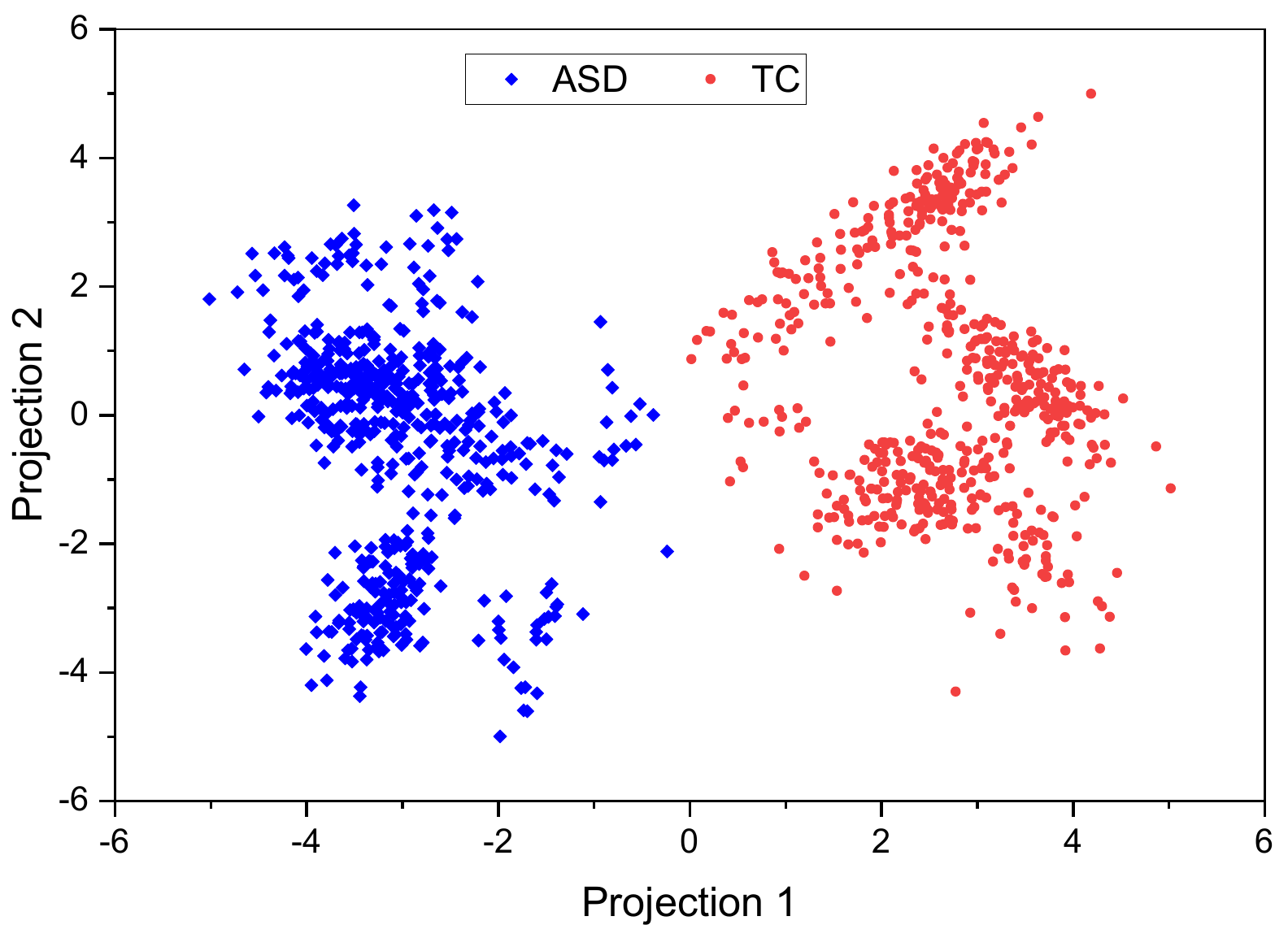}
    \caption{fMRI-TAMCL Full \\
    {\tiny DBI=0.934, CHI=341.7, SC=0.481}
    }
    \label{Figure6e_umap_full}
\end{subfigure}

\caption{Two-dimensional latent space visualization on ABIDE for
different model variants. The projections illustrate the separation between ASD and typically developing control (TC) subjects.}
\label{Figure6_umap_latent_space_variants}

\end{figure*}

% \subsection{Latent Space Visualization and Interpretability}

We apply UMAP (Figure \ref{Figure6_umap_latent_space_variants}) to the ABIDE dataset to assess whether fMRI-TAMCL learns a clinically meaningful latent representation prior to classification. We compare the full model with RTGMFF \cite{jia2025rtgmff} and three ablated variants (NoCL, PairOnly, and NoSup) that remove or modify contrastive objectives governing the shared latent space. The UMAP embeddings show a more progressively structured organization across variants. NoCL yields diffuse, heavily overlapping clusters; PairOnly and RTGMFF \cite{jia2025rtgmff} improve the separation of diagnostic groups; NoSup further tightens clusters; and the full fMRI-TAMCL model produces the clearest, minimally overlapping ASD and TC clusters. Latent space quality metrics support this: fMRI-TAMCL achieves the best clustering, outperforming RTGMFF \cite{jia2025rtgmff} with a lower Davies–Bouldin Index (0.934 vs. 1.387), a higher Calinski-Harabasz index (341.7 vs. 214.8) and a higher Silhouette Coefficient (0.481 vs. 0.341).

\begin{figure*}[!t]
\centering

% ---------- Row 1: three panels ----------
\begin{subfigure}[t]{0.32\textwidth}
    \centering
    \includegraphics[width=\linewidth]{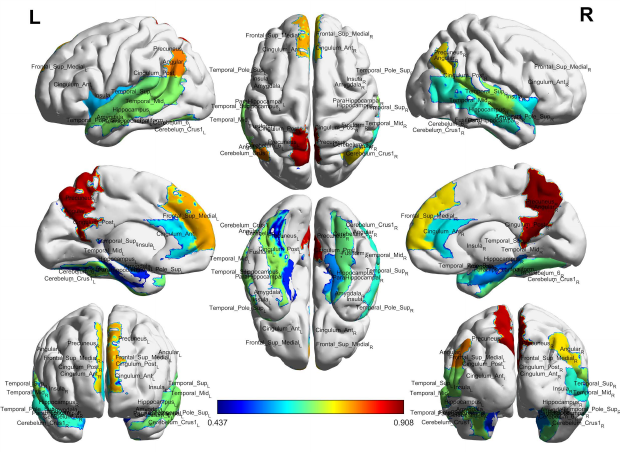}
    \caption{ABIDE (ASD diagnostic group). Top 30 ROIs with $S_r>0.4$.}
    \label{fig:brainnet_abide}
\end{subfigure}
\hfill
\begin{subfigure}[t]{0.32\textwidth}
    \centering
    \includegraphics[width=\linewidth]{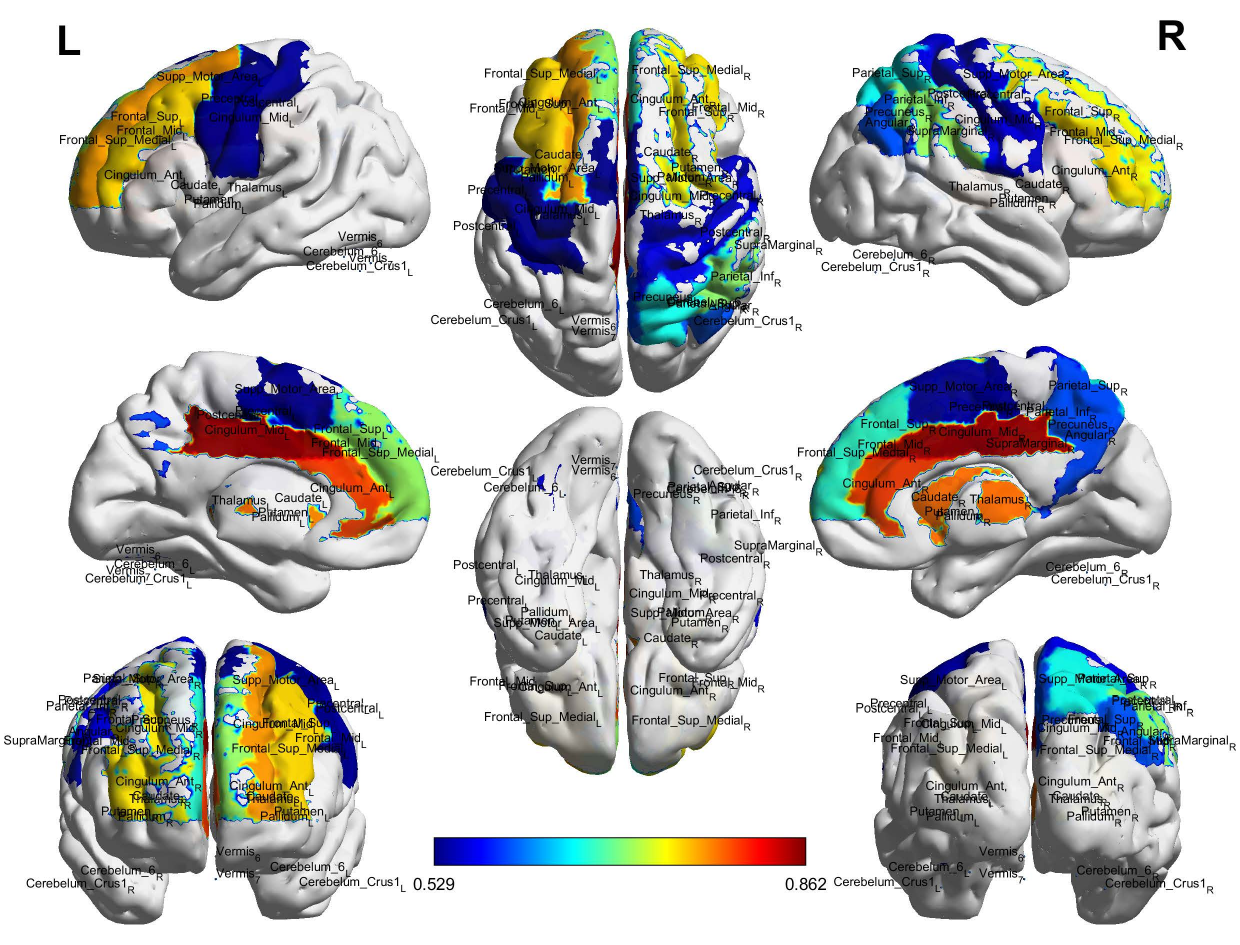}
    \caption{ADHD-200 (ADHD diagnostic group). Top 35 ROIs with $S_r>0.5$.}
    \label{fig:brainnet_adhd}
\end{subfigure}
\hfill
\begin{subfigure}[t]{0.32\textwidth}
    \centering
    \includegraphics[width=\linewidth]{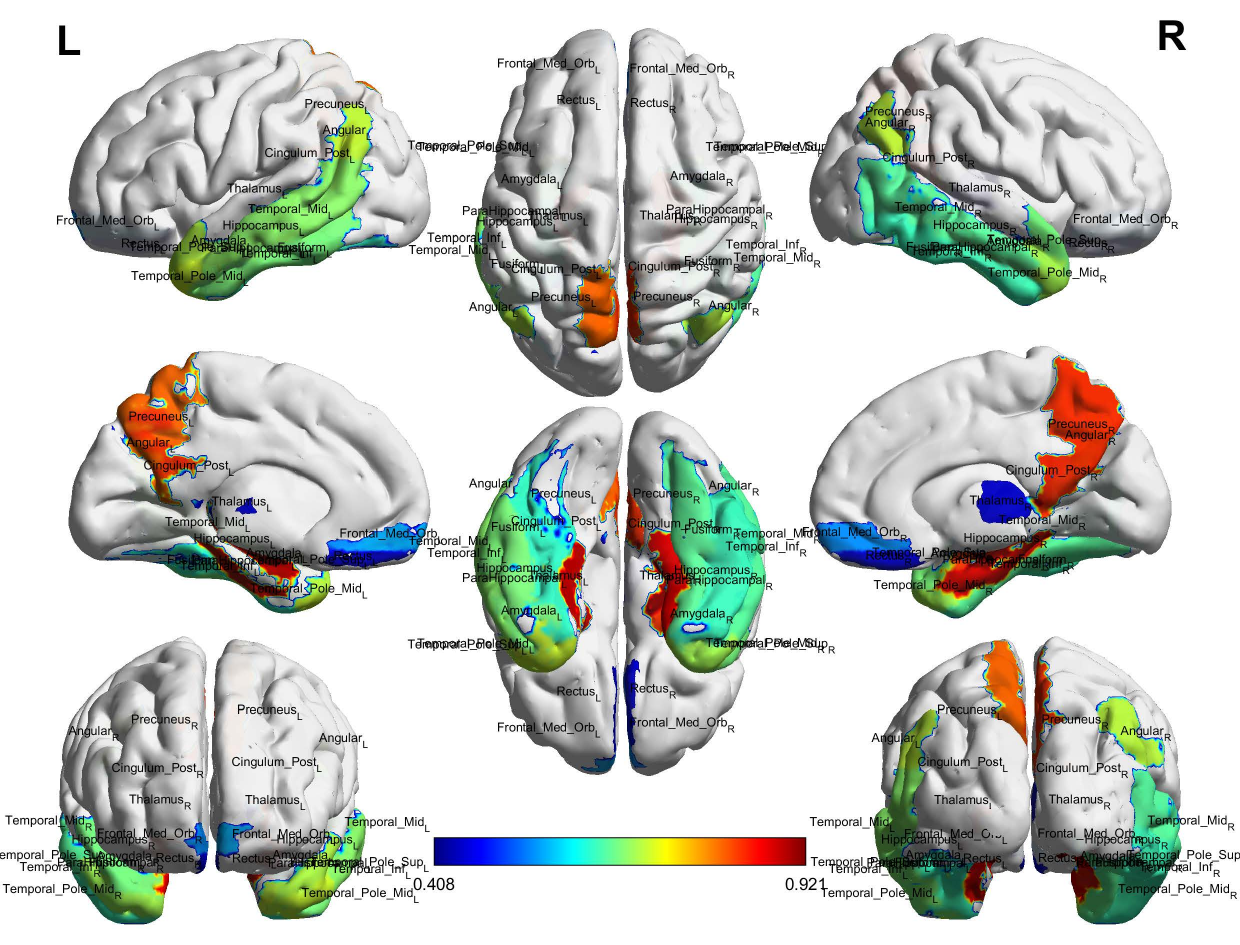}
    \caption{ADNI (AD diagnostic group). Top 28 ROIs with $S_r>0.4$.}
    \label{fig:brainnet_adni}
\end{subfigure}

\vspace{2mm}

% ---------- Row 2: two centered panels ----------
\begin{subfigure}[t]{0.40\textwidth}
    \centering
    \includegraphics[width=\linewidth]{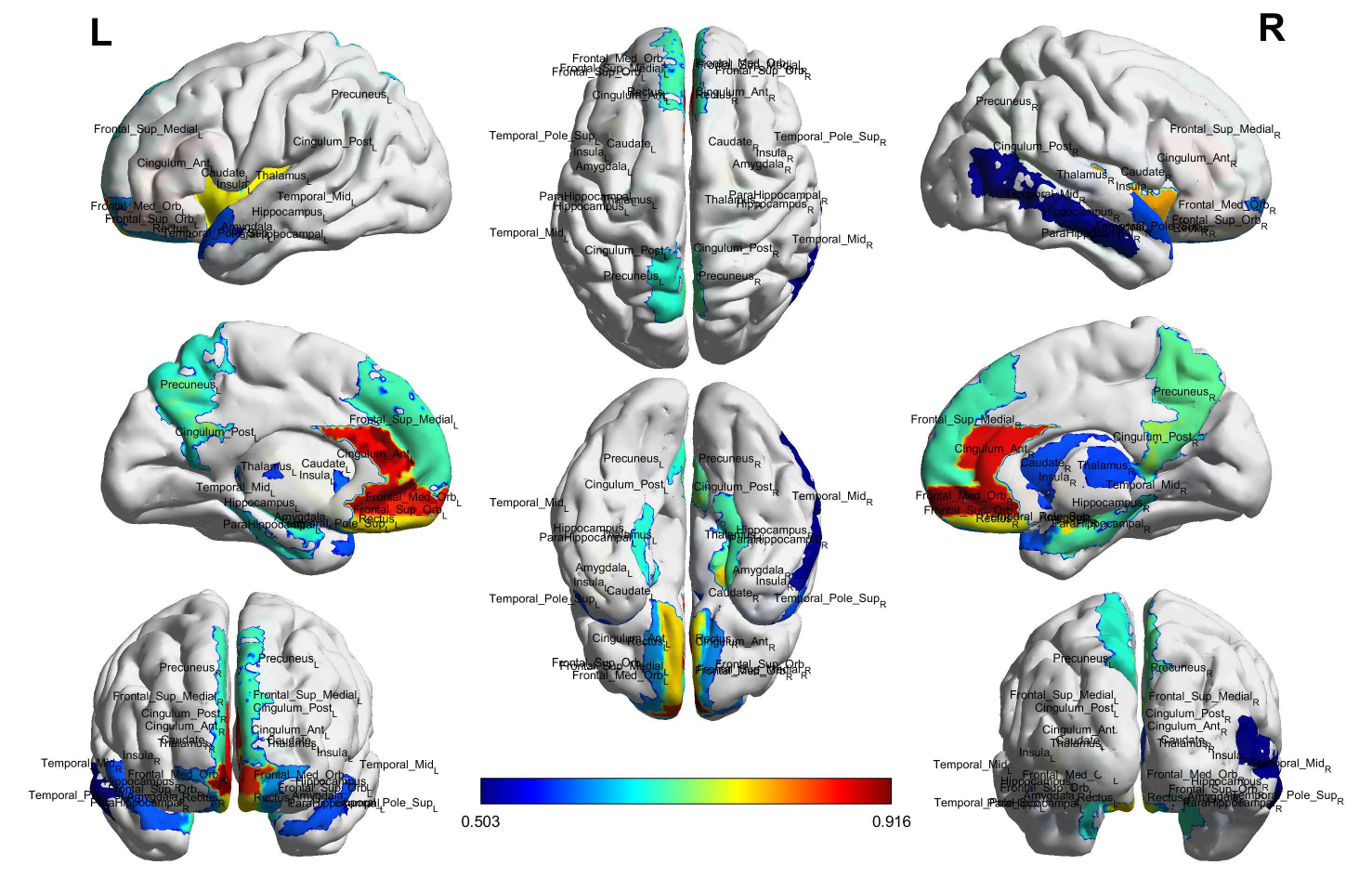}
    \caption{REST-meta-MDD (MDD diagnostic group). Top 30 ROIs with $S_r>0.5$.}
    \label{fig:brainnet_mdd}
\end{subfigure}
\hspace{0.06\textwidth}
\begin{subfigure}[t]{0.40\textwidth}
    \centering
    \includegraphics[width=\linewidth]{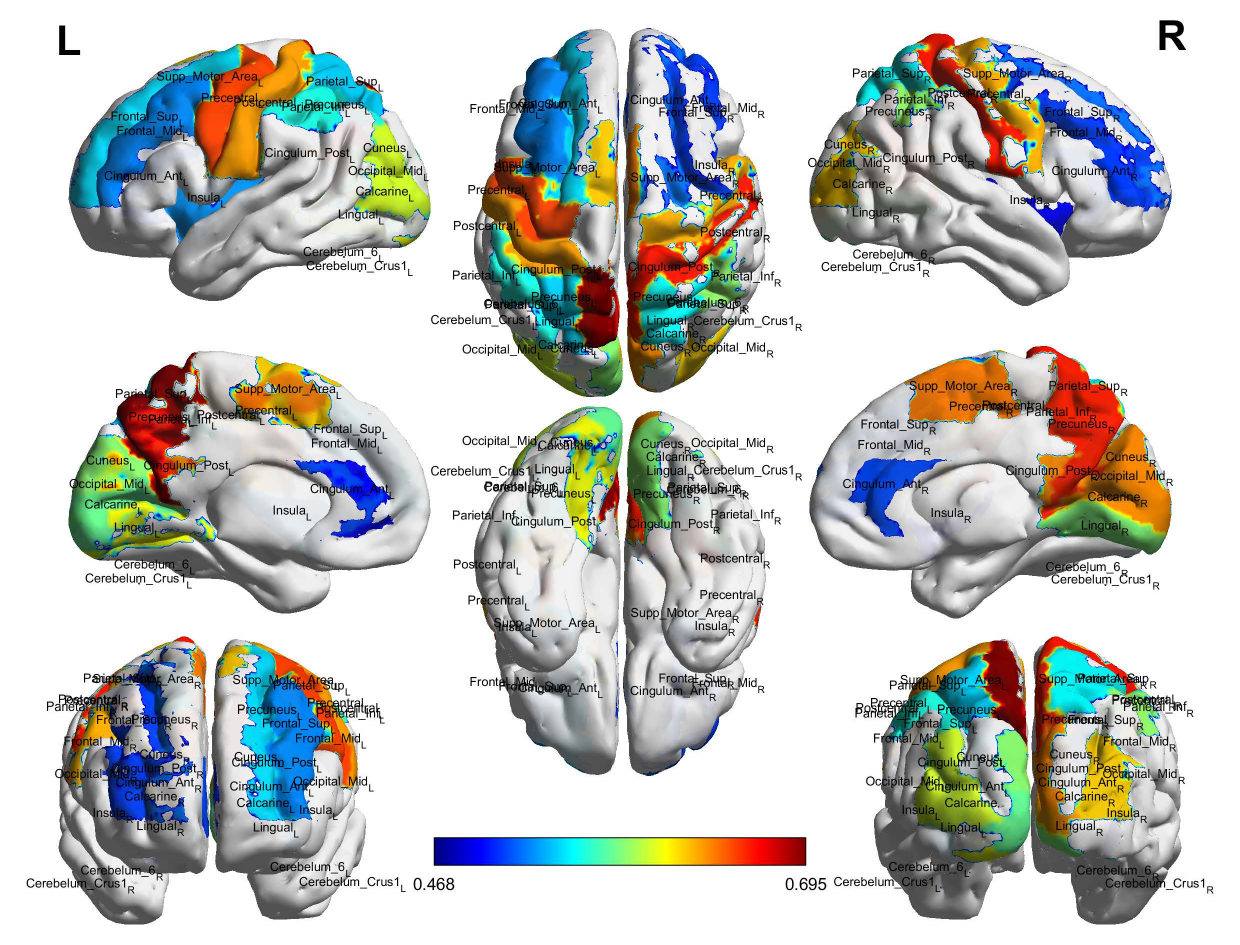}
    \caption{HCP-EP (EP diagnostic group). Top 34 ROIs with $S_r>0.4$.}
    \label{fig:brainnet_hcp_ep}
\end{subfigure}

\caption{Condition-specific brain saliency maps visualized using BrainNet
Viewer on the BrainMesh\_ICBM152\_smoothed surface. The maps show the most
salient ROIs identified for ASD, ADHD, AD, MDD, and EP diagnostic groups.}
\label{figure7_BrainNet_Saliency_Visualization}

\end{figure*}

% \subsection{Biomarker Analysis for Clinical Interpretability}
We performed a clinical interpretability analysis using (1) graph-gradient ROI saliency and (2) cross-dataset consolidated ROI saliency across condition-specific diagnostic groups in five datasets.  Graph-gradient saliency estimates ROI biomarker importance by backpropagating class scores $y_{i,c}$ to the concatenated image, FC, and text embeddings $g_{i,r}$ for each ROI $r$ and subject $i$, yielding MinMax-normalized saliency $S_r$, where
% \[
% S_r = \text{MinMax}\left(\sum_{i} \left| \frac{\partial y_{i,c}}{\partial g_{i,r}} \right| \cdot |g_{i,r}| \right).
% \]
higher values indicate greater class-discriminative importance. ROI saliency maps are projected onto the ICBM152 brain surface with BrainNet Viewer for biomarker visualization.

For ABIDE-ASD (Figure \ref{figure7_BrainNet_Saliency_Visualization}(a)), saliency is dominated by posterior DMN regions, especially Cingulum\_Post\_L (0.908), Precuneus\_R/L (0.906/0.876), and Cingulum\_Post\_R (0.873), with secondary temporal social-cognitive contributions from Temporal\_Sup\_R/L (0.689/0.671), Temporal\_Mid\_L (0.679), and Temporal\_Pole\_Sup\_L (0.673).  
For ADHD-200 (Figure \ref{figure7_BrainNet_Saliency_Visualization}(b)), the most salient biomarkers are cingulo-striatal and cortico-subcortical regions, led by Cingulum\_Mid\_L/R (0.862/0.853), Pallidum\_R/L (0.823/0.813), Putamen\_L (0.804), and Thalamus\_L (0.802), with additional saliency in Cingulum\_Ant\_R/L (0.801/0.796), Caudate\_R/L (0.785/0.776), Frontal\_Sup\_L (0.770), and Frontal\_Mid\_L/R (0.749/0.745), reflecting attention, executive-control, and fronto-striatal involvement.

For ADNI-AD (Figure \ref{figure7_BrainNet_Saliency_Visualization}(c)), the strongest biomarkers are medial temporal-limbic regions: Hippocampus\_R/L (0.921/0.893), ParaHippocampal\_L/R (0.895/0.870), and Amygdala\_R/L (0.845/0.831). This highlights the predominant involvement of the memory-network. Posterior DMN regions also contribute, Cingulum\_Post\_R/L (0.839/0.821), Precuneus\_R/L (0.834/0.806), with additional temporal association saliency (0.632–0.706) suggesting broader temporal-lobe involvement.

For REST-meta-MDD (Figure \ref{figure7_BrainNet_Saliency_Visualization}(d)), saliency concentrates in salience, medial frontal, and affective-limbic regions: Cingulum\_Ant\_L/R (0.916/0.862), Frontal\_Med\_Orb\_R/L (0.912/0.852), Insula\_L/R (0.795/0.762), Amygdala\_R/L (0.773/0.752)-plus Rectus\_R/L (0.776/0.759), hippocampal/parahippocampal regions (0.675–0.698), Cingulum\_Post\_R (0.727), and Precuneus\_R (0.698). This indicates distributed medial prefrontal–limbic and DMN involvement in MDD.

In HCP-EP (Figure \ref{figure7_BrainNet_Saliency_Visualization}(e)), saliency was dominated by distributed sensorimotor, temporal, prefrontal, and subcortical biomarkers, led by Precentral\_R/L (0.835/0.812), Postcentral\_R (0.795), Temporal\_Sup\_R (0.781), Frontal\_Sup\_R (0.768), Thalamus\_R (0.751), and Putamen\_L (0.722), indicating widespread sensorimotor and cognitive-integration abnormalities.

% \subsection{ Cross-Dataset Consolidated ROI Saliency} 

% Figure \ref{Figure8_cross_dataset_biomarker_interpretability} (left panel) shows disorder-specific network biomarkers: ASD includes DMN (0.801); ADHD highlights salience/cingulo-opercular systems (0.813); AD has strongest limbic/medial temporal involvement (0.866); EP shows a distributed network hub pattern (0.668); and MDD is dominated by medial frontal networks (0.850). The right panel shows that the top 20 cross-dataset ROIs are dominated by posterior DMN hubs (bilateral precuneus, posterior cingulate; DMN score 0.597), followed by medial frontal/anterior cingulate (salience/cingulo-opercular; 0.515) and limbic–medial temporal regions (amygdala, hippocampus, parahippocampal; 0.461), particularly in AD and MDD.

\begin{figure}[!t]
\centering

\begin{minipage}[c]{0.35\columnwidth}
    \centering
    \includegraphics[
        width=\linewidth,
        trim=0 0 0 0,
        clip
    ]{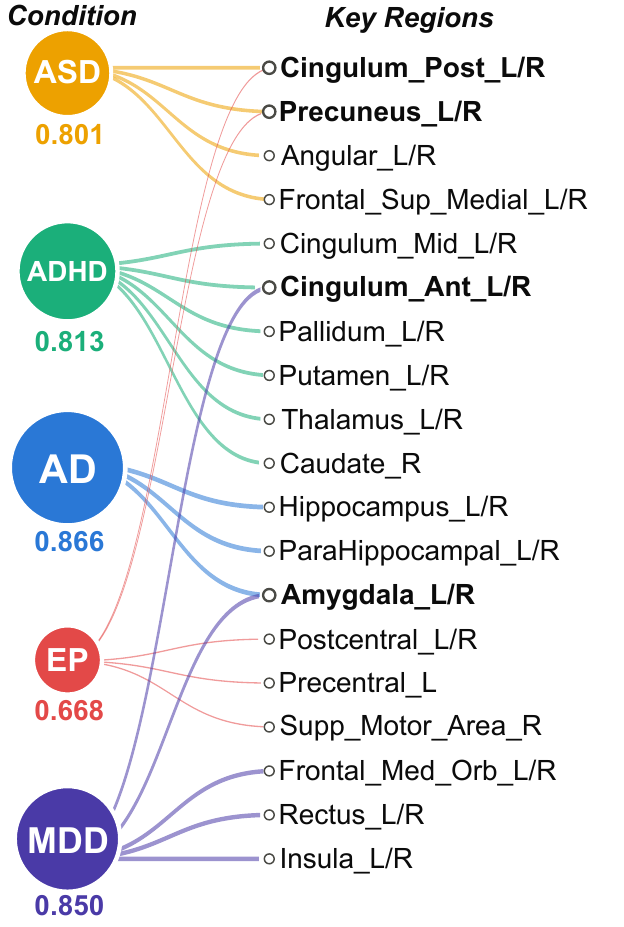}
\end{minipage}
\hspace{0.02\columnwidth}
\begin{minipage}[c]{0.60\columnwidth}
    \centering
    \includegraphics[
        width=\linewidth,
        trim=0 0 0 0,
        clip
    ]{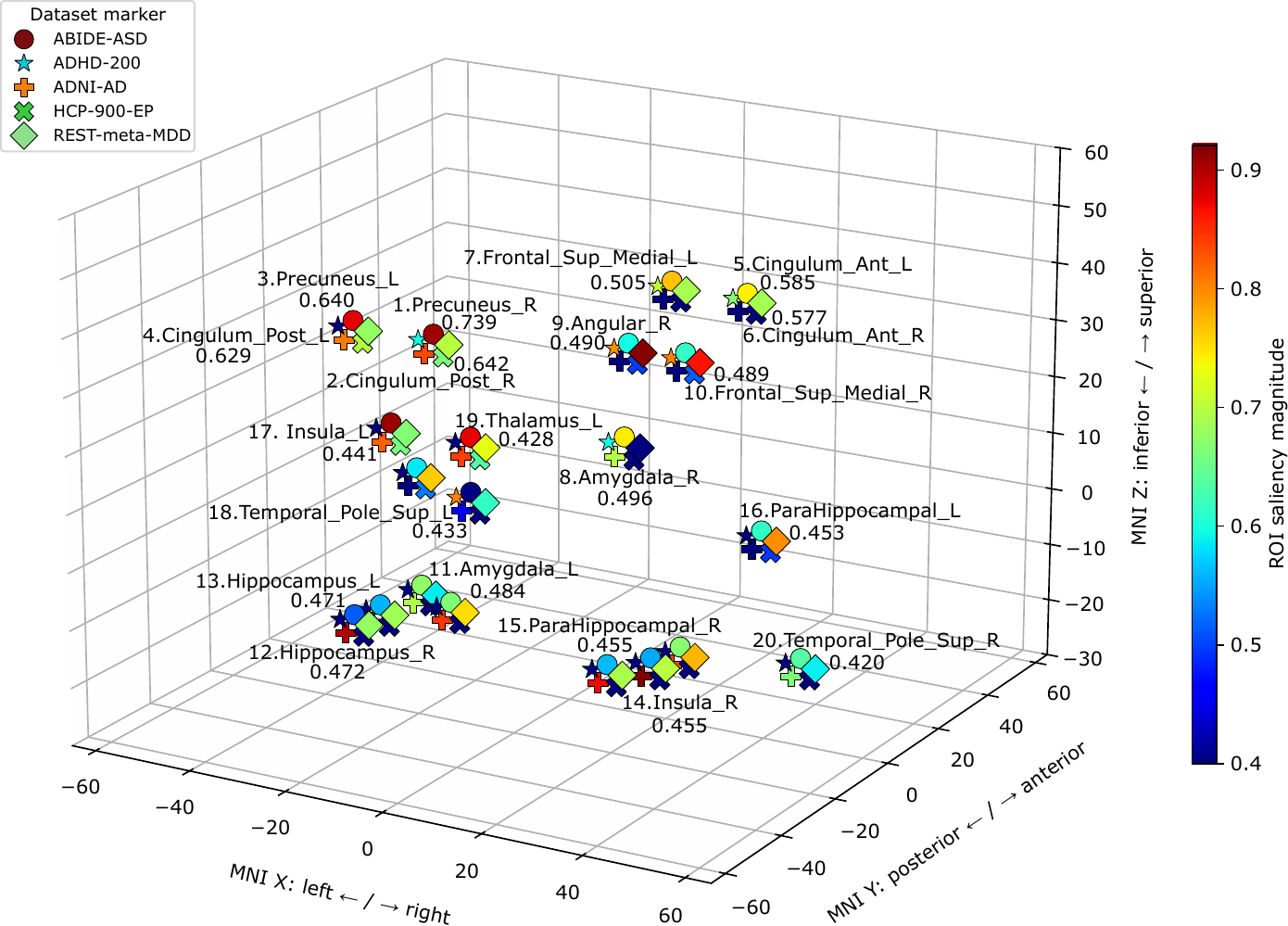}
\end{minipage}

\caption{Cross-dataset biomarker interpretability visualization. Condition–key region mapping (left). ROI saliency in MNI-152 stereotactic space (right).}
\label{Figure8_cross_dataset_biomarker_interpretability}

\end{figure}

Figure \ref{Figure8_cross_dataset_biomarker_interpretability} (left panel) shows disorder-specific network biomarkers: ASD (DMN, 0.801), ADHD (salience/fronto-striatal, 0.813), AD (limbic/medial temporal, 0.866), EP (DMN–sensorimotor, 0.668), and MDD (salience–limbic–medial frontal, 0.850). The right panel shows the top 20 cross-dataset ROIs are dominated by posterior DMN (0.597), followed by medial frontal/anterior cingulate (0.515) and limbic–medial temporal (0.461), with strongest saliency in AD and MDD.

\section{Conclusion}
We introduced fMRI-TAMCL, a text-anchored multimodal contrastive framework that combines fMRI images, sparse FC, and SATD+FVSerial text for brain disorders classification. Across five datasets, it delivers strong diagnostic performance, supported by ablation, robustness, sensitivity, efficiency, latent-space, and biomarker analyzes. SATD improves transferability by replacing fixed activation thresholds with subject-specific BOLD-based thresholds, with the largest gains on HCP-EP, where fixed thresholds failed and saliency revealed a distinct distributed hub-like biomarker pattern. Interpretability analyzes highlight disorder-relevant networks, including DMN in ASD, cingulo-striatal systems in ADHD, medial temporal-limbic regions in AD, and salience–limbic–medial frontal circuitry in MDD. Future work includes hospital-cohort validation, faster inference via pruning, distillation, and lightweight encoders, and improved text refinement using domain-adapted biomedical LLMs.

\section*{References}
\bibliographystyle{IEEEtran}
\bibliography{referencesabv}

% \begin{IEEEbiography}[{\includegraphics[width=1in,height=1.25in,clip,keepaspectratio]{a1.png}}]{First A. Author} (Fellow, IEEE) and all authors may include biographies. Biographies are often not included in conference-related papers. This author is an IEEE Fellow. The first paragraph
% may contain a place and/or date of birth (list place, then date). Next, the author’s educational background is listed. The degrees should be listed with type of degree in what field, which institution,
% city, state, and country, and year the degree was
% earned. The author’s major field of study should
% be lower-cased.
% \end{IEEEbiography}

% \begin{IEEEbiography}[{\includegraphics[width=1in,height=1.25in,clip,keepaspectratio]{a1.png}}]{Second A. Author} (Fellow, IEEE) and all authors may include biographies. Biographies are often not included in conference-related papers. This author is an IEEE Fellow. The first paragraph
% may contain a place and/or date of birth (list place, then date). Next, the author’s educational background is listed. The degrees should be listed with type of degree in what field, which institution,
% city, state, and country, and year the degree was
% earned. The author’s major field of study should
% be lower-cased.
% \end{IEEEbiography}

\end{document}